\documentclass[11pt,a4paper]{article}

\usepackage{xpeng-template}

\usepackage{hyperref}

\usepackage{mathptmx}
\usepackage[scaled=0.92]{helvet}

\usepackage{amsmath}
\usepackage{amssymb}

\usepackage{xpeng-brand}

\usepackage{url}
\usepackage{graphicx}
\usepackage{booktabs}
\usepackage{array}
\usepackage{tabularx}
\usepackage{multirow}
\usepackage{nicefrac}
\usepackage{enumitem}
\usepackage{pifont}
\usepackage{wrapfig}
\usepackage{caption}
\usepackage[numbers,sort&compress]{natbib}
\usepackage{hyperref}
\usepackage{microtype}

\usepackage{algorithm}
\usepackage{algorithmic}

\usepackage{commath}
\usepackage{amsthm}
\usepackage{bm}
\usepackage{graphicx}
\usepackage{subcaption}
\usepackage{booktabs}
\usepackage{adjustbox}
\usepackage{multirow}

\usepackage{mdframed}
\definecolor{promptbg}{RGB}{248,248,248}
\definecolor{promptframe}{RGB}{210,210,210}
\newtcblisting{promptbox}[1][]{
  listing only,
  breakable,
  enhanced,
  colback=promptbg,
  colframe=promptframe,
  arc=2mm,
  boxrule=0.5pt,
  left=1.5mm,
  right=1.5mm,
  top=1mm,
  bottom=1mm,
  listing options={
    basicstyle=\ttfamily\small,
    breaklines=true,
    columns=fullflexible,
    keepspaces=true
  },
  #1
}

\usepackage{listings}
\usepackage{placeins}

\newcommand{\ci}[2]{#1\,{\scriptsize$\pm$\,#2}}

\xpengcorrespondence{Jie Chen}{chenj81@xiaopeng.com}

\title{Engagement Process: Rethinking the Temporal Interface of Action and Observation}

\author{%
  Jialian Li, Yuchen Cao, Junhong Liu, Weiran Guo, Xutao Wang, Jiaming Song, Jiahao Zhang, Jie Chen$^{\dagger}$%
}

\date{}  % leave empty for technical reports

\begin{document}

% --- Page 1: logo, title, authors, abstract panel, optional code/webpage ---
\begin{xpenghero}
  \begin{abstract}
Task completion in digital and physical environments increasingly involves complex temporal interaction, where actions and observations unfold over different time scales rather than align with fixed observation--action steps. To model such interactions, we propose \emph{Engagement Process} (EP), an interaction formalism that inherits the decision-theoretic structure of POMDPs while making time explicit in the action--observation interface. EP represents actions and observations as decoupled event streams along time, rather than updates paired at fixed decision steps. This interface captures single-agent timing issues such as deliberation latency, delayed feedback, and persistent actions, while supporting richer agent-side organization, multi-rate coordination, and compositional interaction among subsystems. Across toy, LLM-agent, and learning experiments, EP exposes temporal behaviors hidden by step-based interfaces and enables policies to adapt under explicit time costs.
\end{abstract}

\end{xpenghero}

% --- Main sections (one file per section; add \input lines as needed) ---
% \input{main/abstract}
\section{Introduction}

Recent advances in large language models\citep{achiam2023gpt,guo2025deepseek}, embodied intelligence\citep{ahn2022can,intelligence2026pi}, and tool-augmented agents\citep{OpenClaw2026,AnthropicClaudeCode2025} have enabled agents to operate in increasingly realistic digital and physical environments. In such settings, interaction is rarely an instantaneous observe--act loop: an agent may spend time reasoning, issue tool calls whose results arrive later, execute actions that persist over time, and receive sensory or communication signals while earlier actions are still unfolding. These timing differences create a basic mismatch between the temporal structure of actions and observations.

One side of this mismatch concerns actions. In step-based formalisms, an action is selected at a decision epoch and applied before the next observation. In realistic interaction, however, producing an action may take time, as in LLM reasoning or tool construction; after initiation, it may persist, as in navigation commands or pending tool calls. These issues have been studied in metareasoning and anytime computation~\citep{russell1991principles,zilberstein1996using,ball2025metareasoning}, as well as in semi-MDPs, options, and hierarchical reinforcement learning~\citep{SuttonPrecupSingh1999Options,hutsebaut2022hierarchical}. The dual side concerns observations. Messages, tool responses, and alarms may arrive sparsely or asynchronously, while sensors such as cameras, audio streams, proprioception, and system logs may produce continuous or multi-rate feedback. Related concerns appear in continuous-time and event-triggered control~\citep{bradtke1994reinforcement,heemels2012introduction,zhang2025overview} and in active perception or sensing~\citep{bajcsy2018revisiting,placed2023survey}. Together, these lines of work show that timing matters on both sides of interaction. What remains less explicit is the joint interface-level problem: actions and observations may have distinct temporal structures, while also overlapping in time. Recent work on real-time reinforcement learning and asynchronous tool-using agents further shows that computation, action execution, and observation arrival can be misaligned in deployed agents~\citep{anokin2025handling,ginart2024asynchronous}.

\begin{figure}
    \centering
    \includegraphics[width=0.9\linewidth]{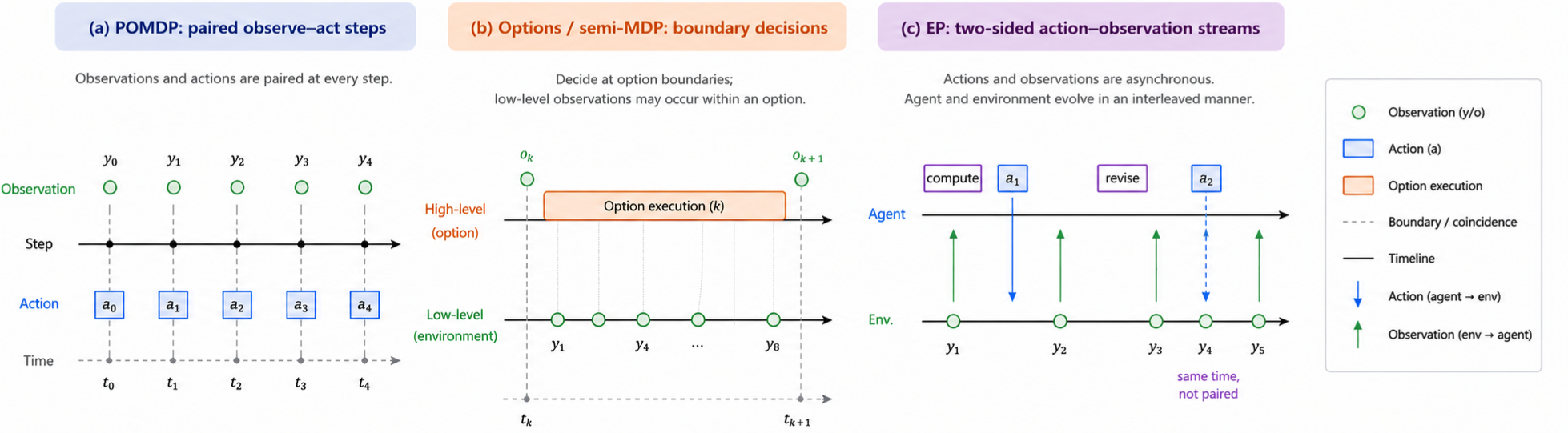}
    \caption{Comparison of interaction interfaces. POMDPs pair observations and actions at fixed decision epochs; options/semi-MDPs organize interaction around temporally extended high-level actions. EP instead separates agent and environment timelines, allowing computation, revision, and multi-rate feedback to be represented as decoupled action--observation events.}
    \label{fig:ep}
    \vspace{-0.2cm}
\end{figure}
This calls for a clean interaction interface in which coupled timing phenomena can be described explicitly. We propose \emph{Engagement Process} (EP), which inherits the decision-theoretic structure of POMDPs while treating actions and observations as decoupled event streams along a shared timeline. At the interface, actions flow from the agent to the environment as timed interventions, while observations flow from the environment to the agent as information events. Actions may take time to compute or persist after initiation, and observations may arrive independently of action completion or decision boundaries. Figure~\ref{fig:ep} contrasts this interface with POMDPs and options/semi-MDPs.

This decoupled interface provides three practical benefits. First, it gives a common language for timing phenomena that otherwise require separate treatments, such as deliberation latency, delayed feedback, persistent actions, interruptions, and multi-rate observations. Second, it separates the external action--observation interface from the agent's internal organization, allowing memory, tool calls, modules, or controllers to operate at different temporal scales. Third, it supports compositional reuse: interactions among subsystems inside an agent can themselves be represented as EP-like processes, with module calls as interventions and returned messages, tool outputs, or sensor updates as observations. By making time explicit at the interface, EP makes temporal structure available to agent modeling and learning rather than leaving it as an implementation detail.

We evaluate EP in three settings. Conceptual toy tasks isolate latency, interruption, and delayed feedback; LLM-based digital and embodied tasks show improved temporal responsiveness in realistic asynchronous workflows; and a training experiment shows that explicit time modeling enables policies to adapt generation to real-time constraints.

Overall, EP does not seek greater state-space expressivity than augmented POMDPs; its contribution is an explicit temporal interface that decouples actions and observations on a shared timeline, providing a unified substrate for heterogeneous interaction, agent-side organization, and compositional subsystem dynamics under explicit time costs.

\section{Related Work}

\paragraph{Decision-process interfaces and temporal abstraction.}
Classical MDPs and POMDPs organize interaction around decision epochs, where observations, actions, and transitions are coupled through a step-based interface~\citep{Puterman1994MDP,KaelblingLittmanCassandra1998POMDP}. SMDPs, options, and hierarchical reinforcement learning relax this structure by introducing variable-duration actions, closed-loop skills, or subtask-level policies~\citep{Howard1971DPS2,SuttonPrecupSingh1999Options,dietterich2000hierarchical}. Metareasoning and anytime computation further study the cost and value of computation before acting~\citep{russell1991principles,zilberstein1996using}. These frameworks capture important temporal and hierarchical structure, but typically organize high-level interaction around decision epochs, action or option intervals, subtask boundaries, or computation policies. EP instead takes the action--observation interface as the primitive: actions and observations are separate time-indexed events, so observations can enter the interface while actions, options, computations, or subtasks are still unfolding.

\paragraph{Continuous-time, event-triggered, and sensing/control systems.}
Continuous-time MDPs and POMDPs move from discrete-time transitions to continuous-time decision dynamics under full or partial observability~\citep{BradtkeDuff1994ContinuousTimeMDP,AltSchultheisKoeppl2020CTPOMDP}. Event-triggered control studies state-dependent execution or update schedules~\citep{Tabuada2007EventTriggered}, hybrid-system models combine continuous dynamics with discrete transitions~\citep{GoebelSanfeliceTeel2012Hybrid}, and active perception or active sensing studies how sensing actions affect information acquisition~\citep{bajcsy2018revisiting}. These frameworks show that timing matters for dynamics, triggers, transitions, and sensing policies. EP is complementary: it does not prescribe a continuous dynamics model, control-triggering rule, or sensing strategy, but defines the agent--environment interaction interface in which agent-initiated actions and state-generated observations appear as decoupled temporal events on a shared timeline.

\paragraph{Streaming, full-duplex, and asynchronous agent systems.}
Recent LLM-agent systems expose practical limitations of strictly serialized interaction. ReAct interleaves reasoning and acting~\citep{YaoEtAl2023ReAct}, Toolformer learns when and how to call external APIs~\citep{SchickEtAl2023Toolformer}, and Reflexion incorporates feedback through verbal reflection~\citep{ShinnEtAl2023Reflexion}, but still largely follow a sequential action--observation loop. Full-duplex, streaming, and asynchronous agents go further by allowing input, output, tool use, and feedback to overlap, while recent benchmarks study interruptions, delayed feedback, asynchronous evidence, and overlapping subtasks~\citep{WangEtAl2024FullDuplex,ma2025language,LinEtAl2025FullDuplexBench,ZhangEtAl2025AViLA,GonzalezPumariegaSuYeanSunkaraChoudhury2025Robotouille,TongWangRenYinWuZhangShen2026StreamingLLM}. EP is motivated by these systems but differs in abstraction level: rather than proposing a new architecture or benchmark, it provides an interaction formalism where computation latency, delayed tool responses, observation-driven revision, and multi-rate feedback share the same action--observation event structure.

\section{Engagement Process}

\subsection{Definition}
We define an \emph{Engagement Process} (EP) over a discrete sequence of ticks
$\mathcal{T}=\{0,1,2,\ldots\}$. Ticks are used as a convenient modeling device:
they specify the resolution at which state evolution, action triggering, and
observation generation are represented. EP is not inherently limited to
discrete time and can be naturally extended to continuous time by replacing ticks
with real-valued event times. EP does not require each tick to contain a
synchronized observation--action pair.

\paragraph{Definition.}
An EP is specified by $\mathcal{E}=(\mathcal{S},\mathcal{A},\mathcal{Y},F,O,U),$, where $\mathcal{S}$ is the state space, $\mathcal{A}$ is the atomic action space, $\mathcal{Y}$ is the observation space, $F$ is the state
transition kernel, $O$ is the observation kernel, and $U$ is the utility function.

\begin{itemize}[leftmargin=*]
    \item \textbf{State.} At tick $t$, the system state $s_t\in\mathcal S$ summarizes the environment, ongoing processes, and interface conditions relevant to future interaction. It may encode persistent action effects, state-dependent observation access, and interaction costs.
    \item \textbf{Actions.} Actions in EP are represented as timed interventions rather than default controls at every tick. The agent chooses an intervention set $A_t\in\mathcal I$, where $\mathcal I\subseteq 2^{\mathcal A}$ may include the empty set or multiple simultaneous atomic actions.
    \item \textbf{Dynamics.} Given $s_t$ and newly triggered interventions $A_t$, the next state evolves as $s_{t+1}\sim F(\cdot\mid s_t,A_t)$. Since $s_t$ may encode ongoing or pending processes, the system can continue to evolve even when no new intervention is triggered, i.e., when $A_t=\emptyset$.
    \item \textbf{Observations.} Observations are generated from the state process, rather than being tied to decision steps. Let $Y_t\subseteq\mathcal{Y}$ denote the observation event set at tick $t$, and let $O(\cdot\mid s_t)$ be a kernel over such event sets, so that $Y_t\sim O(\cdot\mid s_t)$. The set may be empty or contain multiple observations. Because $s_t$ may encode information-access conditions, $O$ can express state-dependent access to information while remaining state-conditioned. Importantly, $Y_t$ and $A_t$ need not be paired.
    \item \textbf{Utility.} Instantaneous utility is generated as $u_t\sim U(\cdot\mid s_t,A_t)$, which can represent task rewards as well as action, sensing, interruption, and deliberation-time costs. When such quantities are encoded in the state, this reduces to $u_t\sim U(\cdot\mid s_t)$. For a per-tick discount factor
$\gamma\in(0,1]$, the cumulative utility over horizon $T$ is $J=\mathbb{E}\left[\sum_{t=0}^{T-1}\gamma^t u_t\right].$
\end{itemize}

\subsection{Relation to standard decision interfaces}

A standard POMDP-style loop can be recovered as a synchronized case of EP. Each tick is a decision epoch, a new observation is consumed before action selection, a single instantaneous action is issued, and the system transitions to the next tick. Under this restriction, actions and observations are paired by construction, and the usual belief-state policy and value-function machinery apply.

EP keeps this decision-theoretic compatibility but changes the interaction primitive. While an augmented POMDP can encode pending processes in the state, EP exposes such timing mechanisms at the interface by allowing actions and observations to appear as separate temporal events. Thus belief states, value functions, and Bellman recursions can still be defined once an information state and utility model are specified, but the interface on which they operate can include delayed computation, persistent interventions, empty ticks, asynchronous observations, and multi-rate feedback.

\subsection{Information state and policy}

Since the system state $s_t$ may not be directly observable, the agent acts from its tick-indexed interaction history $h_t=(A_0,Y_0,u_0,\ldots,A_{t-1},Y_{t-1},u_{t-1}),$
where each intervention set $A_k$ and observation event set $Y_k$ may be empty. We write the agent's information state as
$w_t=\phi(h_t,t),$
where $\phi$ may be a Bayesian belief state, full history, recurrent memory, learned embedding, or a richer agent-side state that tracks pending computations, tool calls, module states, or urgency.

A policy over EP maps this information state to a distribution over admissible intervention sets,
$\pi(\cdot\mid w_t)\in\Delta(\mathcal I),\ A_t\sim\pi(\cdot\mid w_t),$
where $\mathcal I\subseteq 2^{\mathcal A}$ is the intervention-set space. Thus an EP policy may trigger no intervention, one intervention, or multiple simultaneous interventions at a tick. The environment then evolves and generates observations according to
$s_{t+1}\sim F(\cdot\mid s_t,A_t),\ Y_{t+1}\sim O(\cdot\mid s_{t+1}).$
The new observation events are appended to the history to form $w_{t+1}$. This separates the agent's internal decision process from external observation and intervention timing. Detailed belief-state updates, value functions, Bellman recursions, and the reductions to standard POMDPs and SMDPs/options are provided in Appendix~\ref{app:decision_ep}.

\section{Agent-Side Organization and Compositionality}
\label{sec:compositionality}

EP is not an agent architecture, but its interface leaves room for richer agent-side organization than a single observe--act loop. We highlight useful consequences for complex LLM and embodied agents.

\paragraph{Richer agent-side state.}
Because actions and observations are not forced to be paired at fixed decision steps, the agent can maintain richer internal state while computations, tool calls, or pending processes unfold over time. These processes can be represented in the information state $w_t$, which may track memory, outstanding tool calls, module states, deadlines, or urgency. For example, a computer-use assistant may continue reasoning while waiting for a browser action or tool result, and revise its next intervention when a new page state arrives. Thus EP separates the external interaction interface from the agent's internal organization for deciding when and how to intervene.

\paragraph{Coordination across temporal scales.}
This separation is useful for agents whose components operate at different frequencies or latencies. A high-level planner may update slowly, a low-level controller may act frequently, a perception module may stream observations, and a tool call may return after an unpredictable delay. In an embodied robot, for instance, a language-level planner may reason over seconds, a controller may update at high frequency, and cameras or tactile sensors may produce continuous or multi-rate feedback. Under EP, these processes can be placed on the same temporal interface: external actions, tool calls, controller commands, and internal module requests are interventions, while returned tool results, module messages, controller states, and sensor updates are observation events. This view does not require every component to act at every tick; it allows heterogeneous operations to be scheduled and observed through a common event-based interface.

\paragraph{Compositional reuse of the interface.}
The same interface can also be reused inside the agent. A subsystem $M^i$ can be viewed as an EP-like interaction process with its own local interface. Its local interventions $A^i_t$ may include calls to lower-level controllers, tool invocations, module requests, or actions sent to the external environment. Its local observations $Y^i_t$ may include instructions from higher-level modules, messages from peer modules, sensor updates, tool results, or returns from lower-level components. For example, an embodied ``cognitive'' module can treat a navigation controller, manipulation controller, memory module, and perception stack as interacting subsystems, each exchanging interventions and observations through the same abstract interface. In this way, a hierarchical agent can be described by recursively applying the same action--observation event abstraction at different levels, rather than introducing a separate interface for each module boundary.

In this sense, EP provides a homogeneous interface across levels: agent--environment interaction and module--module interaction can be described using the same action--observation event abstraction. This compositional view does not require a particular hierarchy or scheduler, but makes hierarchical, asynchronous, and multi-rate agent designs easier to specify within a single formal language.

\section{Experiments}

We evaluate EP through toy, LLM-based, and learning experiments. Toy experiments isolate deliberation latency, temporally extended execution, and interruption; the LLM-based experiments test asynchronous messages, delayed tool responses, and coordination among agent-side processes operating on a shared timeline; and the learning experiment studies policy adaptation when deliberation time is explicitly modeled. Together, these experiments test whether EP exposes temporal structure that fixed step-based formulations tend to hide or handle through external patches.

% 1. subtables
% 2. test-time scaling citation
% 3. gif snapshot
\subsection{Toy Experiments}
\label{sec:toy-experiments}
We first evaluate EP in controlled toy environments that isolate deliberation latency and temporally extended execution with asynchronous interruption. Using tabular Q-learning or hand-written patches, these experiments test whether the interface exposes temporal structure needed for appropriate decisions rather than benchmark performance. Detailed configurations and results are in Appendix~\ref{app:toy-experiment-configs}.

\subsubsection{Urgency-Aware Deliberation}
\label{sec:urgency-aware-deliberation}
The first experiments study action-generation latency, framing test-time scaling \cite{wei2022chain,snell2024scaling,brown2024large} as a decision on how much computation to allocate before acting, with EP charging this computation against deadline slack. Each task includes a deadline, reward, timeout/failure penalty, and difficulty level. The agent selects one of 5 deliberation modes prior to acting; higher modes are more accurate but slower. \textbf{EP} and \textbf{Step} environments are identical except that EP advances global time during deliberation, whereas Step treats deliberation as instantaneous.

For difficulty $u$, mode $m$ succeeds with probability $\Pr(\mathrm{success}\mid m,u)=\sigma(\alpha_m-\beta u)$, where $\alpha_m$ increases with mode quality and $\beta=3.5$. Thus, the experiments isolate how learning changes when accuracy-improving computation consumes real time.

\paragraph{Single-task setting.}
The single-task experiment is a minimal contextual decision problem where each episode contains one sampled task with urgency and difficulty as inputs. We train a tabular Q-learning agent in EP and Step environments and evaluate both policies in both settings.
\begin{table}[t]
\caption{Urgency-aware deliberation under real-time EP evaluation: in both single-task and sequential settings, EP-trained policy avoids the timeout-prone overuse of slowest mode learned by step training.}
\label{tab:toy1}
\centering
\small
\begin{tabular}{llrrrr}
\hline
\textbf{Setting} & \textbf{Training $\rightarrow$ EP eval.} & \textbf{Mean return $\uparrow$} & \textbf{Success rate $\uparrow$} & \textbf{Timeout rate $\downarrow$} & \textbf{Mode 5 usage} \\
\hline
\multirow{2}{*}{Single-task} & EP $\rightarrow$ EP & $\textbf{+0.800}$ & $\textbf{46.7\%}$ & $\textbf{0.0\%}$ & $21.2\%$ \\
& Step $\rightarrow$ EP & $-1.032$ & $16.1\%$ & $79.0\%$ & $93.6\%$ \\
\hline
\multirow{2}{*}{Sequential-task} & EP $\rightarrow$ EP & $\textbf{+10.07}$ & $\textbf{50.1\%}$ & $\textbf{3.7\%}$ & $15.9\%$ \\
& Step $\rightarrow$ EP & $-14.98$ & $8.4\%$ & $88.9\%$ & $84.0\%$ \\
\hline
\end{tabular}
\vspace{-0.1cm}
\end{table}

The single-task rows of Table~\ref{tab:toy1} highlight a key failure of the step abstraction: since deeper reasoning improves accuracy without time cost during training, the step-trained agent selects mode 5 in over $93\%$ of episodes, leading to $79.0\%$ timeouts and negative return under EP evaluation. In contrast, the EP-trained agent uses mode 5 only $21.2\%$ of the time, preserving deadline slack and achieving higher return, higher success, and no timeouts when computation consumes real time.

\paragraph{Sequential-task setting.}
The second deliberation experiment applies the same mechanism to a fixed task chain with absolute deadlines, where accumulated deliberation time can cause later tasks to expire. It evaluates the computation–urgency tradeoff under interacting deadlines. 

The sequential-task rows of Table~\ref{tab:toy1} show that the sequential setting amplifies this effect: early over-deliberation exhausts the shared budget and causes later task failures. The step-trained policy times out in $88.9\%$ of EP evaluations with mean return $-14.98$, while EP learns to reserve deep modes for slack states, reducing mode 5 usage to $15.9\%$ and improving mean return to $+10.07$.

\paragraph{Toy 1 summary.}
The deliberation experiments show that making reasoning instantaneous distorts the learned policy: the step abstraction overuses deep reasoning, while EP adapts computation depth to urgency and downstream deadlines.

\subsubsection{Hierarchical Patrol with Interruptions}
\label{sec:hierarchical-patrol-interruptions}
The second experiment studies temporally extended execution with a two-level controller: the upper level decides whether to continue a checkpoint routine or respond to an alarm, while the lower level handles navigation. Asynchronous alarms at distant checkpoints yield large rewards if resolved and penalties if missed. This setting instantiates the compositional view in Section~\ref{sec:compositionality}, where lower-level modules act as EP-like subsystems and upper-level commands intervene, with phase progress and completion signals feeding back as observations that can override ongoing execution.

\paragraph{Module-level interruption.}
In the first patrol experiment, checkpoint handling is a single module. Both agents share the same dynamics and primitives, but \textbf{EP} can interrupt navigation or handling at any tick, while the \textbf{Loop} baseline can decide only after the current module completes.
\begin{table}[t]
\caption{Hierarchical patrol with interruptions: fixed boundaries at module and state levels introduce response latency, while EP makes routines observable and interruptible, improving alarm handling and performance over loop-based and patched methods.}
\label{tab:toy2}
\centering
\small
\begin{subtable}{\linewidth}
\caption{Module-level interruption.}
\label{tab:toy2a}
\centering
\begin{tabular}{lrrrr}
\hline
\textbf{Method} & \textbf{Mean return $\uparrow$} & \textbf{Resolve rate $\uparrow$} & \textbf{Expire rate $\downarrow$} & \textbf{Ticks per alarm $\downarrow$} \\
\hline
EP & $\textbf{+896.6}$ & $\textbf{90.9\%}$ & $\textbf{8.0\%}$ & $\textbf{10.9}$ \\
Loop & $-163.5$ & $51.5\%$ & $47.1\%$ & $14.4$ \\
\hline
\end{tabular}
\end{subtable}
\vspace{-0.3cm}
\begin{subtable}{\linewidth}
\caption{State-level interruption.}
\label{tab:toy2b}
\centering
\begin{tabular}{llrrrr}
\hline
\textbf{Depth} & \textbf{Method} & \textbf{Mean return $\uparrow$} & \textbf{Resolve rate $\uparrow$} & \textbf{Expire rate $\downarrow$} & \textbf{Interrupt cost} \\
\hline
2 & EP & $\textbf{+175.1}$ & $\textbf{62.8\%}$ & $\textbf{36.5\%}$ & $61.7$ \\
2 & PatchPro & $+25.1$ & $53.9\%$ & $45.4\%$ & $28.0$ \\
2 & Patch & $-156.5$ & $47.6\%$ & $51.6\%$ & $63.7$ \\
2 & Loop & $-345.0$ & $34.9\%$ & $64.4\%$ & $0.0$ \\
\hline
3 & EP & $\textbf{+143.3}$ & $\textbf{62.4\%}$ & $\textbf{36.8\%}$ & $80.6$ \\
3 & PatchPro & $+68.8$ & $57.2\%$ & $42.0\%$ & $48.4$ \\
3 & Patch & $-101.0$ & $48.4\%$ & $50.8\%$ & $27.1$ \\
3 & Loop & $-298.7$ & $37.5\%$ & $61.7\%$ & $0.0$ \\
\hline
\end{tabular}
\end{subtable}
\vspace{-0.1cm}
\end{table}
Table~\ref{tab:toy2a} shows that EP resolves $90.9\%$ of alarms, compared to $51.5\%$ for Loop. Although both use the same alarm-response primitive, Loop cannot interrupt checkpoint handling, causing many alarms to expire due to comparable handling times and deadlines. EP avoids this limitation by allowing interruptions during execution.

\paragraph{State-level interruption.}
This experiment splits checkpoint handling into phases with durations and interruption costs. \textbf{EP} learns a continue-or-interrupt policy from alarm urgency, current module or phase, remaining phase time, and distance. It is compared with a phase-boundary \textbf{Loop} baseline, a threshold-based \textbf{Patch}, and a stronger \textbf{PatchPro} that also considers reachability, finishability, phase progress, and interruption cost. Detailed rules appear in Appendix~\ref{app:toy-experiment-configs}. Table~\ref{tab:toy2b} shows the same ranking in both phase structures: EP performs best, followed by PatchPro, Patch, and Loop. Loop avoids interruption costs but misses urgent alarms due to lack of within-phase interruption. Patch reacts faster but ignores interactions among distance, deadlines, progress, and costs. PatchPro reduces the gap via hand-designed heuristics, but still underperforms EP.

These results show the benefit of making interruption part of the interaction process rather than an external patch. In EP, an ongoing routine has observable state and can be evaluated or superseded, allowing the policy to learn tradeoffs among urgency, distance, phase progress, and interruption cost. Rule-based patches require more cases and retuned thresholds as the environment changes.

\paragraph{Toy 2 summary.}
The patrol experiments show that EP enables more flexible temporal execution. By making routines observable and interruptible, it allows urgent alarms to affect decisions before phase completion, improving alarm resolution and return without hand-crafted interruption rules.

\subsection{LLM-based Experiments}

\begin{figure}[t]
\centering
% \resizebox{0.95\linewidth}{0.85\height}{
%   \includegraphics{figs/6_2_experiment.png}
% }
\includegraphics[
    width=0.95\linewidth,
    height=0.35\textheight
]{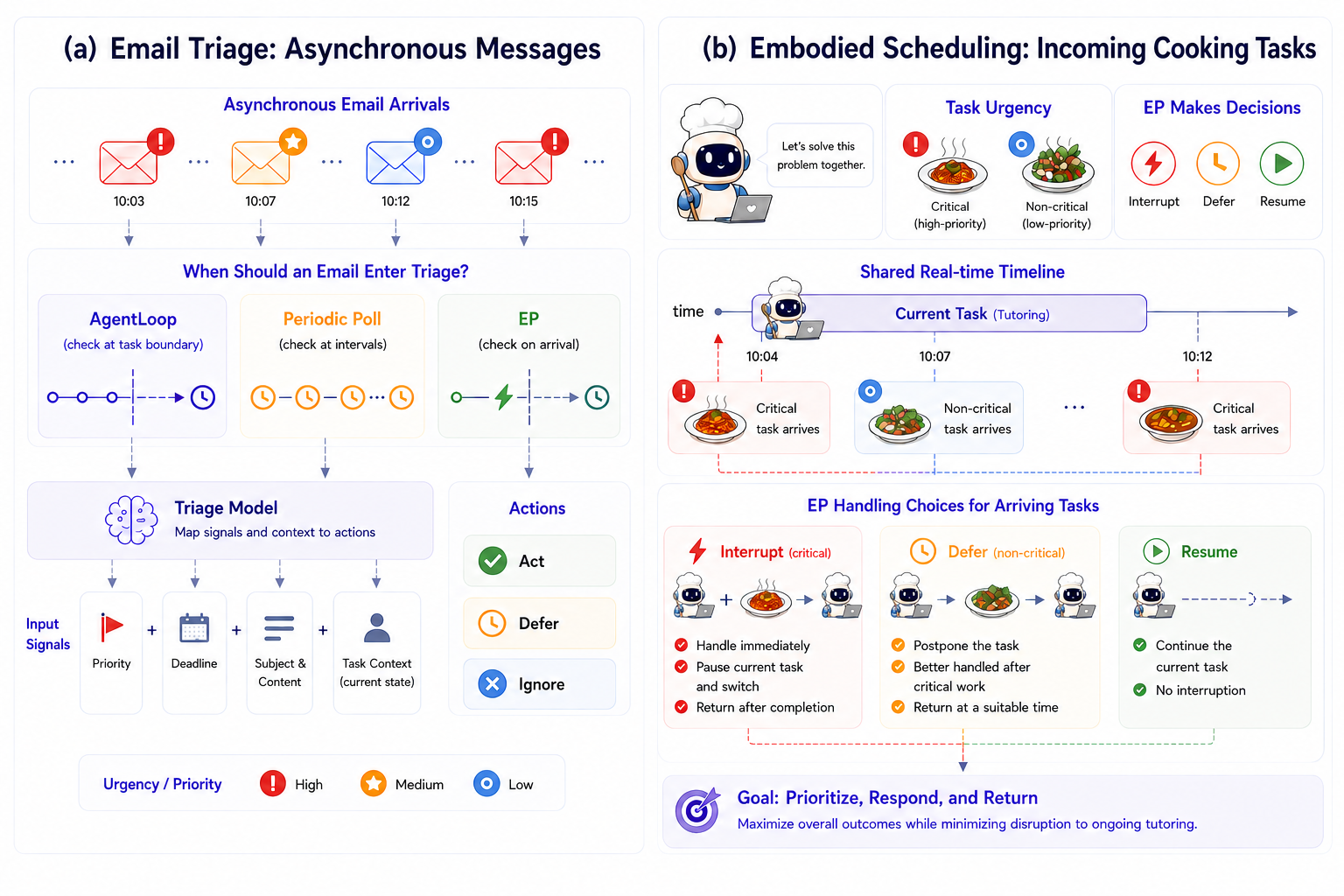}
\vspace{-0.2cm}
\caption{LLM-based experiments. Tasks can be interpreted as a triage and scheduling problem over a shared timeline. EP exposes ongoing work and incoming
events in a way that lets the controller decide whether to interrupt, defer, or
continue.}
\label{fig:llm_experiment}
\vspace{-0.22cm}
\end{figure}

We now move from toy environments to LLM-driven workflows. Figure~\ref{fig:llm_experiment} illustrates two case studies: a digital assistant responding to asynchronous emails while continuing a main task, and an embodied controller coordinating tutoring with evolving cooking processes. In both, generation, monitoring, feedback, and scheduling unfold on a shared timeline: generated tokens proxy elapsed ticks for LLM-side work, while environment states continue to evolve in the background.We compare EP with common agent-loop and periodic-polling interfaces, focusing on whether a decoupled action--observation interface better supports such asynchronous LLM workflows.

\subsubsection{Digital Assistant with Asynchronous Emails}

\label{sec:digital_assistant_async_emails}
This experiment evaluates whether a digital assistant can make progress on a long workplace task while responding to time-sensitive asynchronous emails. Each episode starts with one main task, such as a delivery plan, root-cause analysis, or launch-readiness document. Main-task execution is realized as LLM generation and consumes simulated time. During generation, emails arrive stochastically; each email has an urgency level and a deadline. The assistant may check the inbox, open a message, handle it, set a reminder, or resume the main task.

The \textbf{Milestones} setting uses milestone-level subtasks, while the \textbf{Single} setting uses a single longer main-task unit. All methods use the same main-task outputs, email streams, triage model, action durations, and scoring rule. They differ in how observations are coupled to ongoing generation. \textbf{AgentLoop} follows a synchronized loop: new emails are observed only after the current main-task unit finishes. \textbf{Periodic Poll} adds fixed-interval polling during generation, snapshots visible emails at polling points, handles them, and then resumes the main task. \textbf{EP} treats main-task generation as an interruptible timed action: email arrivals and reminders are delivered as events while generation is unfolding, allowing the agent to interrupt, triage, act, and later resume partially completed output.

% We report DeepSeek-V3.2 results in the main text to keep the table compact; Qwen3.5-Plus results and configuration details are provided in Appendix~\ref{app:digital-assistant-config}. For each main-task decomposition, \emph{single} or \emph{milestones}, we use 200 seeds and 5 episodes per seed for each method, yielding 1000 episodes per method.
We report DeepSeek-V3.2 results in the main text, while Qwen3.5-Plus results and configuration details are provided in Appendix~\ref{app:digital-assistant-config}. For each decomposition setting (\emph{single} or \emph{milestones}), we use 200 seeds and 5 episodes per seed, yielding 1000 episodes per method.

\begin{table}[t]
\centering
\small
\caption{Digital assistant with asynchronous emails using DeepSeek-V3.2 as the email-triage model. \emph{Utility} is the total reward under this rescore setup; \emph{Balanced} is the combined main-task/email score defined above; \emph{Timeout} is the fraction of emails that miss their deadline; \emph{Latency} is the average first-response delay to emails; and \emph{Main} is the normalized main-task completion score.}
\label{tab:digital_assistant_async_email}
\begin{adjustbox}{width=0.8\linewidth}
\begin{tabular}{llrrrrr}
\toprule
\textbf{Setting} & \textbf{Method} &
\textbf{Utility} &
\textbf{Balanced} &
\textbf{Timeout} &
\textbf{Latency} &
\textbf{Main} \\
\midrule
Single     & EP    & \textbf{3.023} & \textbf{0.792} & \textbf{0.315} &  \textbf{6.92} & 0.984 \\
Single     & Periodic Poll & 2.016 & 0.619 & 0.426 & 10.15 & 0.983 \\
Single     & AgentLoop  & 0.730 & 0.382 & 0.598 & 13.58 & \textbf{1.000} \\
\midrule
Milestones & EP    & \textbf{3.113} & \textbf{0.818} & \textbf{0.305} &  \textbf{6.91} & 0.995 \\
Milestones & Periodic Poll & 1.775 & 0.606 & 0.444 & 11.33 & 0.998 \\
Milestones & AgentLoop  & 1.591 & 0.569 & 0.467 & 11.83 & \textbf{1.000} \\
\bottomrule
\end{tabular}
\end{adjustbox}
\vspace{-0.2cm}
\end{table}

% \begin{figure}
%     \centering
%     \includegraphics[width=0.9\linewidth]{figs/digital_assistant_async_emails_core_comparison.png}
%     \caption{Overview of the Digital Assistant with Asynchronous Emails.}
%     \label{fig:digital_assistant_with_asynchronous_emails_overview}
% \end{figure}

\paragraph{Results.}
Table~\ref{tab:digital_assistant_async_email} shows that EP obtains the highest utility and balanced score in both %DeepSeek-V3.2 
settings, with Periodic Poll consistently between EP and AgentLoop. The main gain comes from responsiveness. Averaged over \emph{single} and \emph{milestones}, EP reduces timeout rate from 0.533 for AgentLoop to 0.310, while Periodic Poll reaches 0.435. EP also reduces first-response latency from 12.71 to 6.92, compared with 10.74 for Periodic Poll.

The aggregate scores reflect the same pattern. Relative to AgentLoop, EP improves utility by 2.293 points in \emph{single} and 1.522 points in \emph{milestones}; the corresponding balanced-score gains are 0.410 and 0.249. AgentLoop attains the highest main-task score because it rarely interrupts generation, but this comes at the cost of many missed email deadlines. Polling mitigates the failure but remains tied to an external checking schedule. EP instead exposes email arrivals as events at their arrival times, producing a better tradeoff between main-task progress and time-sensitive email handling.

% TODO:
% - tool latency
% - async messages
% - 三种baseline对比

\subsubsection{Embodied Scheduling Task}
\label{sec:embodied_scheduling}

The second LLM-based experiment is an embodied scheduling task coupling timing with physical state. The controller coordinates tutoring and independently evolving cooking processes, with some actions feasible only in compatible states. A shared timeline is therefore essential for aligning generation, movement, dish evolution, and readiness signals. 
We evaluate \textbf{one-dish} and \textbf{multi-dish} variants: the former isolates reaction latency, while the latter adds overlapping deadlines and tutoring work. All successfully completed tutoring tasks and dishes yield an identical base reward, while elapsed ticks, task switching, and polling operations incur penalties (see Appendix~\ref{app:embodied-scheduling}).

% \paragraph{Interaction interfaces.}
All methods share the same dishes, tutoring streams, action space, time model, and utility; they differ only in when cooking events enter the controller context and who decides the response. \textbf{AgentLoop} treats tutoring as an atomic step and surfaces cooking signals only after the current tutoring problem completes. \textbf{Periodic Poll} checks for newly fired signals at fixed intervals during tutoring. This bounds delay and resembles a common design in coding agents, where a long-running background shell command is monitored through periodic status checks rather than event-triggered interruption. However, visibility is still tied to a manually chosen polling cadence: a signal that fires just after a poll remains hidden until the next poll, even as the agent continues to pay checking overhead. \textbf{ReflexPatch} exposes signals immediately but first applies a default engine-side kitchen response before returning control. \textbf{EP} exposes the fired signal at the interruption point and lets the controller decide whether to react, defer, or resume tutoring.

\begin{table}[t]
\centering
\caption{
Embodied scheduling results using DeepSeek-V3.2. Utility is the total episode score; Dish/Crit.
reports dish success in the one-dish setting and critical-dish success in
the multi-dish setting. Soft qual. reports soft-dish quality in
multi-dish episodes.
}
\label{tab:embodied_scheduling}
\begin{adjustbox}{width=0.9\linewidth}
\begin{tabular}{llccccc}
\toprule
\textbf{Setting} & \textbf{Method} & \textbf{Utility} $\uparrow$ & \textbf{Dish/Crit.} $\uparrow$ & \textbf{Soft qual.} $\uparrow$ & \textbf{Vis. delay} $\downarrow$ & \textbf{Finish delay} $\downarrow$ \\
\midrule
One-dish & AgentLoop & 3.604 & 0.266 & -- & 224.8 & 227.8 \\
One-dish & Periodic Poll & 5.598 & 0.922 & -- & 4.7 & 17.2 \\
One-dish & ReflexPatch & 5.254 & 0.938 & -- & 4.5 & 16.0 \\
One-dish & EP & \textbf{5.731} & \textbf{0.938} & -- & \textbf{3.1} & \textbf{15.5} \\
\midrule
Multi-dish & AgentLoop & 19.671 & 0.745 & 0.833 & 74.1 & 91.7 \\
Multi-dish & Periodic Poll & 18.981 & 0.718 & \textbf{0.971} & 18.0 & 37.6 \\
Multi-dish & ReflexPatch & 21.202 & 0.913 & 0.952 & 17.9 & 28.6 \\
Multi-dish & EP & \textbf{21.288} & \textbf{0.936} & 0.957 & \textbf{9.7} & \textbf{28.5} \\
\bottomrule
\end{tabular}
\end{adjustbox}
\vspace{-0.2cm}
\end{table}

\paragraph{Results.}
Table~\ref{tab:embodied_scheduling} shows one-dish failures mainly reflect
delayed visibility. AgentLoop often observes readiness only after tutoring ends;
Periodic Poll reduces this delay with in-tutoring checks; ReflexPatch and EP
benefit from immediate signal exposure. The small ReflexPatch--EP gap suggests this low-conflict setting mostly probes signal timing rather than
scheduling arbitration. Multi-dish episodes introduce overlapping deadlines and tutoring work. Periodic Poll lowers average delay, but a fixed cadence can still pay repeated checking
costs while missing signals until the next poll. This is often tolerable for
soft dishes with keep-warm slack, but less reliable for critical hard-window
dishes. 

ReflexPatch also performs well via engine-side urgent responses. Thus, the distinction lies less in raw performance than in interface structure: EP unifies tutoring, evolution, signaling, and resumption within a common event timeline rather than relying on polling schedules or external reflex rules. The modest ReflexPatch--EP gap may partly reflect that current pretrained LLM policies are not yet optimized for event-driven temporal reasoning.

This embodied task operationalizes the compositional framework of Section~\ref{sec:compositionality}: generation, environment dynamics, and scheduling are unified under a common action--observation interface. With tutor tokens serving as temporal ticks, we next examine whether RL can induce time awareness through rewards for both correctness and urgency-appropriate generation length.

\subsection{Time-aware LLM training}
\label{sec:time_aware_llm_training}

We next test whether EP can guide reasoning models to account for deliberation time. Here, generating tokens consumes time, and each math problem is paired with a requested token budget
$B\in\{1000,2000,4000,8000\}$
and a natural-language urgency instruction. The goal is to answer correctly while adapting the amount of reasoning to the available budget.

\paragraph{Setup and baselines.}
We train three GRPO variants from the same Qwen3-8B thinking model~\citep{yang2025qwen3} on difficult DeepMath-103K data~\citep{DeepMath103K}. \textbf{EP-aware} gives reward 1 only when the answer is correct and the generated solution stays within the requested budget. \textbf{RLVR-8K} uses the same 8000-token limit but rewards only correctness, while \textbf{RLVR-16K} uses a 16000-token limit and the same correctness-only reward. Thus EP-aware versus RLVR-8K isolates the budgeted reward, while RLVR-8K versus RLVR-16K tests whether a shorter generation cap alone induces time-aware behavior. All models are evaluated with a 32768-token decoding limit; training and evaluation details are in Appendix~\ref{app:time_aware_llm_training}.

\begin{table}[t]
\centering
\small
\caption{Time-aware LLM training results. Budgeted
Success requires both correctness and satisfying the requested token budget. We
show the most urgent and least urgent budgets.}
\label{tab:time_aware_llm_main}
\begin{tabular}{lrrrrr}
\hline
\textbf{Model} & \textbf{BS@1K} & \textbf{BS@8K} & \textbf{Acc@1K} & \textbf{Tok@1K} & \textbf{Tok@8K} \\
\hline
Qwen3-8B & 0.000 & 0.228 & 0.407 & 7554 & 8977 \\
RLVR-16K & 0.037 & 0.669 & 0.782 & 5120 & 5599 \\
RLVR-8K & 0.284 & \textbf{0.795} & \textbf{0.786} & 1892 & 2413 \\
EP-aware & \textbf{0.733} & 0.784 & 0.765 & \textbf{421} & \textbf{1409} \\
\hline
\end{tabular}
\vspace{-0.2cm}
\end{table}

\paragraph{Results.}
Table~\ref{tab:time_aware_llm_main} shows that correctness-only RL improves raw accuracy but does not reliably produce budget-aware behavior. Under the 1000-token budget, RLVR-8K reaches 0.786 Accuracy but only 0.284 Budgeted Success, whereas EP-aware training reaches 0.733 Budgeted Success with much shorter generations. The full results are shown in Table~\ref{tab:time_aware_llm_training}. The comparison between RLVR-8K and RLVR-16K shows that shorter training caps have some implicit effect, but much less than directly optimizing the EP-aware budgeted objective. EP-aware training produces the shortest outputs under tight budgets and increases output length when more budget is available, indicating that the model learns a time-aware generation policy rather than merely benefiting from a shorter decoding cap.
\section{Conclusion and Discussion}
We introduced \emph{Engagement Process} (EP), a formalism for representing the temporal interface between action and observation. EP models actions as timed interventions and observations as state-generated information events over a shared evolving state, making temporally decoupled interaction explicit rather than forcing it into a synchronized step loop. Across toy environments, LLM-based workflows, and time-aware training, our experiments show that this interface exposes temporal structure needed for responsive execution, interruption handling, and adaptation to explicit time costs.

EP is best viewed as an interaction interface that complements standard decision-theoretic models. Its goal is not to replace augmented-state representations, but to expose timing structure that such representations often hide inside state and transition design. As an interface formalism, EP leaves agent memory, subsystem scheduling, and policy learning to architectures and algorithms built on top of it. Future work includes EP-aware architectures and learning algorithms, richer continuous-time formulations, and larger-scale embodied, LLM-based, hierarchical, and multi-agent evaluations.

% --- References ---
\bibliographystyle{unsrtnat}
\bibliography{refs}

% --- Appendix (optional; comment out \appendix block if unused) ---
\appendix
\appendix
\newpage
% \section{Appendix}
\section{Extended Related Work}\label{app:related_work}                                                           
\paragraph{Streaming, full-duplex, and asynchronous agent systems.}
Recent agent systems increasingly expose interaction patterns that go beyond turn-based loops. Full-duplex spoken dialogue models allow agents to listen while speaking so that input and output overlap in time~\citep{WangEtAl2024FullDuplex,ma2025language,LinEtAl2025FullDuplexBench}. Streaming multimodal agents extend this pattern to vision and other modalities, processing continuous sensory input alongside action generation~\citep{ZhangEtAl2025AViLA}. 

A complementary line of work studies real-time reasoning in dynamic environments. Wen et al.~\citep{WenYeZhangYangZhu2025RealTimeReasoning} introduce Real-Time Reasoning Gym and AgileThinker, which combine reactive and planning-style reasoning to balance reasoning depth against response latency. Tong et al.~\citep{TongWangRenYinWuZhangShen2026StreamingLLM} provide a broader survey of streaming LLM systems and propose a taxonomy spanning output-streaming, sequential-streaming, and concurrent-streaming architectures.

Latency-awareness has also become increasingly important in embodied AI. Zheng et al.~\citep{ZhengMaoZhangCai2025RRARA} propose the Time Conversion Mechanism (TCM), which converts inference latency into simulation-time delay, together with an asynchronous reflex-and-reflection architecture for real-time embodied decision making. Gonzalez-Pumariega et al.~\citep{GonzalezPumariegaSuYeanSunkaraChoudhury2025Robotouille} introduce Robotouille, a benchmark for asynchronous long-horizon embodied planning with interruptions, overlapping subtasks, and delayed feedback.

These systems collectively highlight the growing importance of latency-aware interaction and asynchronous feedback in modern agent systems.

\paragraph{Active sensing and information gathering.}
In many systems, the availability of observations depends on the agent’s sensing, communication, or attention state. This connects EP to active sensing and information-gathering work, where agents select sensing actions to acquire useful information~\citep{VeigaRenoux2023ActiveSensingSurvey}. EP differs in scope: rather than focusing specifically on sensing policies, it provides a general interaction formalism in which sensing actions, delayed observations, multi-rate signals, and ordinary task actions can all coexist within the same temporally decoupled interface.

\section{Decision-Theoretic Details for EP}
\label{app:decision_ep}

EP defines an interaction interface in which actions and observations are
represented as temporally situated events. This appendix shows how standard
decision-theoretic objects---history, information state, belief, value, and
Bellman recursion---can be defined on top of this interface. The main change from
a standard step-based formulation is that the policy chooses an intervention set
$A_t\in\mathcal I$, and the observation model returns an observation event set
$Y_t\subseteq\mathcal Y$.

\subsection{History, information state, and policy}

At tick $t$, the agent has access to its tick-indexed interaction history
$$
h_t=(A_0,Y_1,u_0,\ldots,A_{t-1},Y_t,u_{t-1}),
$$
where $A_k\in\mathcal I$ is the intervention set chosen at tick $k$, $Y_{k+1}$ is
the observation event set received after the transition to tick $k+1$, and
$u_k$ is the utility generated at tick $k$. Either $A_k$ or $Y_{k+1}$ may be
empty. This indexing matches the EP transition order
$$
s_{t+1}\sim F(\cdot\mid s_t,A_t),\qquad
Y_{t+1}\sim O(\cdot\mid s_{t+1}).
$$

An information state is any agent-maintained representation
$$
w_t=\phi(h_t,t).
$$
It may be the full history, a Bayesian belief state, a recurrent memory, or a
learned embedding. In richer agent systems, $w_t$ may also track pending
computations, outstanding tool calls, module states, deadlines, or urgency.

A policy over EP maps the information state to a distribution over admissible
intervention sets:
$$
\pi(\cdot\mid w_t)\in\Delta(\mathcal I),
\qquad
A_t\sim\pi(\cdot\mid w_t),
$$
where $\mathcal I\subseteq 2^{\mathcal A}$ is the intervention-set space. Thus an
EP policy may choose no intervention, a single intervention, or multiple
simultaneous interventions at a tick.

\subsection{Belief update}

A belief state is an idealized information state that retains the posterior over
the latent system state:
$$
b_t(s)=\mathbb P(s_t=s\mid h_t,t).
$$
The explicit dependence on $t$ allows the belief to change with elapsed time,
even if no new observation event arrives.

Given belief $b_t$ and intervention set $A_t$, the predictive belief before
observing $Y_{t+1}$ is
$$
\bar b_{t+1}(s')
=
\sum_{s\in\mathcal S}
F(s'\mid s,A_t)b_t(s).
$$
After receiving the observation event set $Y_{t+1}$, the posterior is
$$
b_{t+1}(s')
=
\eta\,
O(Y_{t+1}\mid s')\,
\bar b_{t+1}(s'),
$$
where $\eta$ normalizes over $s'\in\mathcal S$. The predictive distribution over
observation event sets is
$$
P(Y_{t+1}\mid b_t,A_t)
=
\sum_{s'\in\mathcal S}
O(Y_{t+1}\mid s')\bar b_{t+1}(s').
$$
This formulation also covers empty observation sets: receiving no event,
$Y_{t+1}=\emptyset$, can still be informative when event absence is
state-dependent.

Persistent actions, delayed effects, pending tool calls, or ongoing processes
can be represented by including the relevant variables in $s_t$. Their evolution
is then handled by the transition kernel $F(s'\mid s,A_t)$, rather than by
introducing a separate duration variable.

\subsection{Value functions and Bellman recursion}

Given an information state $w_t$, the value of policy $\pi$ over horizon $T$ is
$$
V_t^\pi(w_t)
=
\mathbb E_\pi
\left[
\sum_{k=t}^{T-1}\gamma^{k-t}u_k
\,\middle|\, w_t
\right].
$$
When the belief state is sufficient for prediction and control, we write
$V_t^\pi(b_t)$ and $\pi(\cdot\mid b_t)$.

Let
$$
r(b_t,A_t)=\mathbb E[u_t\mid b_t,A_t]
$$
denote the expected immediate utility, marginalizing over the next state and
observation event set under $F$ and $O$. The policy value satisfies
$$
V_t^\pi(b_t)
=
\mathbb E_{A_t\sim\pi(\cdot\mid b_t)}
\left[
r(b_t,A_t)
+
\gamma\,
\mathbb E_{Y_{t+1}\sim P(\cdot\mid b_t,A_t)}
\left[
V_{t+1}^\pi(b_{t+1})
\right]
\right],
$$
where $b_{t+1}$ is obtained by the EP belief update above. The optimal value
satisfies
$$
V_t^*(b_t)
=
\max_{A_t\in\mathcal I}
\left[
r(b_t,A_t)
+
\gamma\,
\mathbb E_{Y_{t+1}\sim P(\cdot\mid b_t,A_t)}
\left[
V_{t+1}^*(b_{t+1})
\right]
\right].
$$
Thus the Bellman structure carries over to EP, with the standard action space
replaced by the intervention-set space and the standard observation model
replaced by an event-set observation kernel.

\subsection{Recovering standard POMDPs}

A standard POMDP is recovered as a synchronized special case of EP. Restrict the
intervention-set space so that each set contains exactly one action,
$A_t=\{a_t\}$, assume that actions are instantaneous, and require each tick to
produce exactly one observation event, $Y_{t+1}=\{y_{t+1}\}$. If
$$
F(s'\mid s,A_t)=P(s'\mid s,a_t),
\qquad
O(\{y_{t+1}\}\mid s')=P(y_{t+1}\mid s'),
$$
then the EP belief update becomes
$$
b_{t+1}(s')
=
\eta\,
P(y_{t+1}\mid s')
\sum_{s\in\mathcal S}
P(s'\mid s,a_t)b_t(s),
$$
which is the usual POMDP belief update. Under this synchronized restriction,
observations and actions are paired by construction, and standard POMDP policies,
value functions, and Bellman recursions are recovered.

\subsection{Encoding SMDPs and options}

SMDP and option-style interaction can also be represented as restricted EPs. An
option can be encoded by allowing an intervention $o\in\mathcal A$ to create an
``active option'' component in the state. While this component is active, the
transition kernel $F$ evolves both the environment and the option's internal
closed-loop execution. Termination can be represented as a state-dependent flag
or event, after which the high-level policy is allowed to initiate another
option.

During option execution, the admissible intervention sets can be restricted so
that the high-level policy either issues no new option or follows the option's
internal execution rule until termination. This recovers the standard
boundary-organized structure of options and SMDPs. EP differs by not requiring
termination to be the only point at which observations enter the high-level
interface: observation event sets may arrive while an option, action, or other
process is still unfolding, and the information state can be updated before that
process terminates.

Thus POMDPs, SMDPs, and options can be viewed as synchronized or structured
restrictions of EP. EP keeps their decision-theoretic machinery but relaxes the
requirement that actions, observations, and high-level decisions share the same
boundary.

% Include this file after \appendix in the main paper.
% \newcommand{\ci}[2]{#1\,{\scriptsize$\pm$\,#2}}
\section{Toy Experiment Details and Additional Results}
\label{app:toy-experiment-configs}

This appendix collects the numerical settings used in the toy experiments in
Section~\ref{sec:toy-experiments}. The main text focuses on the interaction
mechanism being tested and the resulting behavior; the tables below provide the
concrete values needed to reproduce each setting. Within each experiment, EP and
the corresponding baselines share the same task distribution, rewards, penalties,
low-level dynamics, and available primitives unless the decision interface is
explicitly listed as the object of comparison.

\paragraph{Deliberation experiments.}
Tables~\ref{tab:toy1a-config} and~\ref{tab:toy1b-config} specify the two
urgency-aware deliberation settings. The mode qualities $\alpha_m$ are used in
the success model described in Section~\ref{sec:urgency-aware-deliberation};
the associated mode durations are counted against the clock only in the EP
formulation. Thus, the two formulations share the same quality profile and
differ only in whether quality-improving computation consumes time.

\begin{table}[t]
\centering
\small
\begin{tabular}{@{}p{0.28\linewidth}p{0.64\linewidth}@{}}
\hline
\textbf{Parameter} & \textbf{Value} \\
\hline
Tasks per episode & $1$ \\
Slack choices & $\{0.3, 1.0, 2.0, 4.0, 8.0\}$ seconds \\
Difficulty & $u \sim \mathrm{Uniform}(0,1)$; bin edges at $0.33$ and $0.66$ \\
Mode durations & $\{0.2, 0.8, 1.6, 3.0, 5.0\}$ seconds \\
Mode qualities $\alpha_m$ & $\{0.0, 0.8, 1.6, 2.4, 3.2\}$ \\
Task reward and penalty & $+4$ for success; $-2$ for failure or timeout \\
Training and evaluation & $8{,}000$ training episodes; $3{,}000$ evaluation episodes; seed $42$ \\
\hline
\end{tabular}
\caption{Configuration for the single-task urgency-aware deliberation experiment.}
\label{tab:toy1a-config}
\end{table}

\begin{table}[t]
\caption{Configuration for the sequential urgency-aware deliberation experiment.}
\label{tab:toy1b-config}
\centering
\small
\begin{tabular}{@{}p{0.28\linewidth}p{0.64\linewidth}@{}}
\hline
\textbf{Parameter} & \textbf{Value} \\
\hline
Tasks per episode & $10$ \\
Deadline gaps & $\{0.4, 0.9, 1.6, 3.5, 5.5\}$ seconds between adjacent tasks \\
Difficulty & $u \sim \mathrm{Uniform}(0,1)$; bin edges at $0.33$ and $0.66$ \\
Mode durations & $\{0.2, 0.8, 1.6, 3.0, 5.0\}$ seconds \\
Mode qualities $\alpha_m$ & $\{0.0, 0.8, 1.6, 2.4, 3.2\}$ \\
Task reward and penalty & $+4$ for success; $-2$ for failure or timeout \\
Discounting & $\gamma_{\mathrm{second}}=0.995$; $\gamma_{\mathrm{task}}=0.95$ \\
Training and evaluation & $8{,}000$ training episodes; $3{,}000$ evaluation episodes; seed $42$ \\
\hline
\end{tabular}
\end{table}

\begin{figure}[t]
\centering
\includegraphics[width=\linewidth]{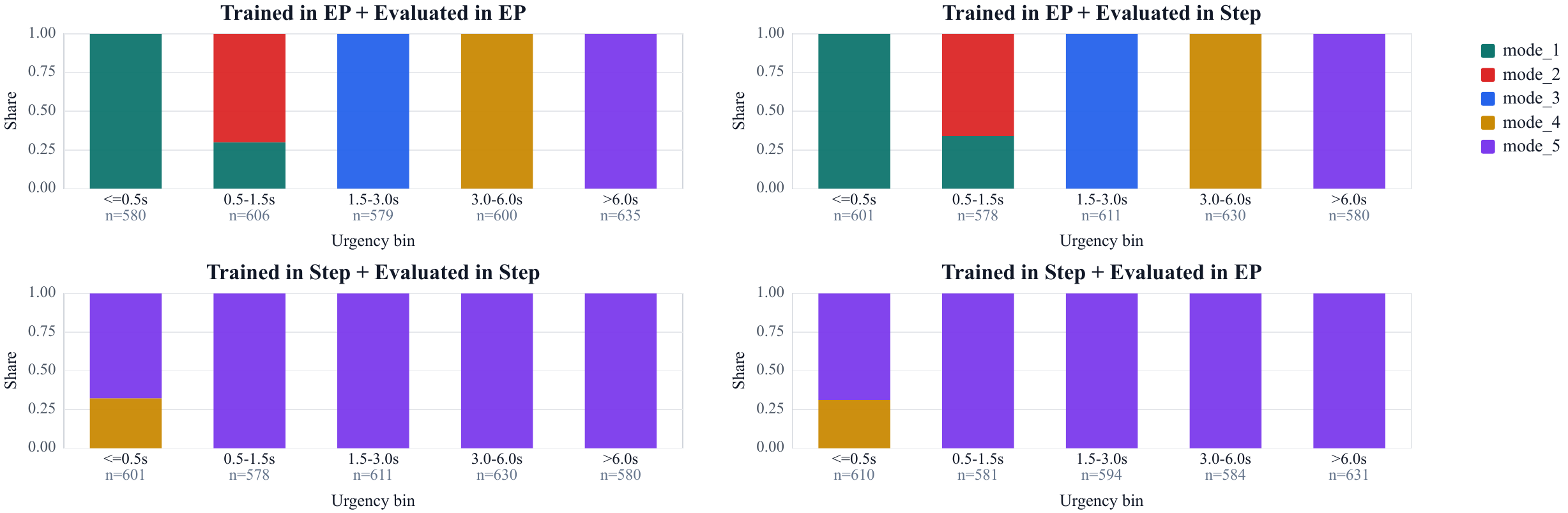}
\caption{Urgency-conditioned deliberation-mode distributions in the single-task setting. The EP-trained policy spreads probability mass across modes according to remaining slack, whereas the step-trained policy heavily favors the deepest mode because deliberation is free during training.}
\label{fig:toy1a-urgency-mode-distribution}
\end{figure}

\begin{figure}[htbp]
\centering
\includegraphics[width=\linewidth]{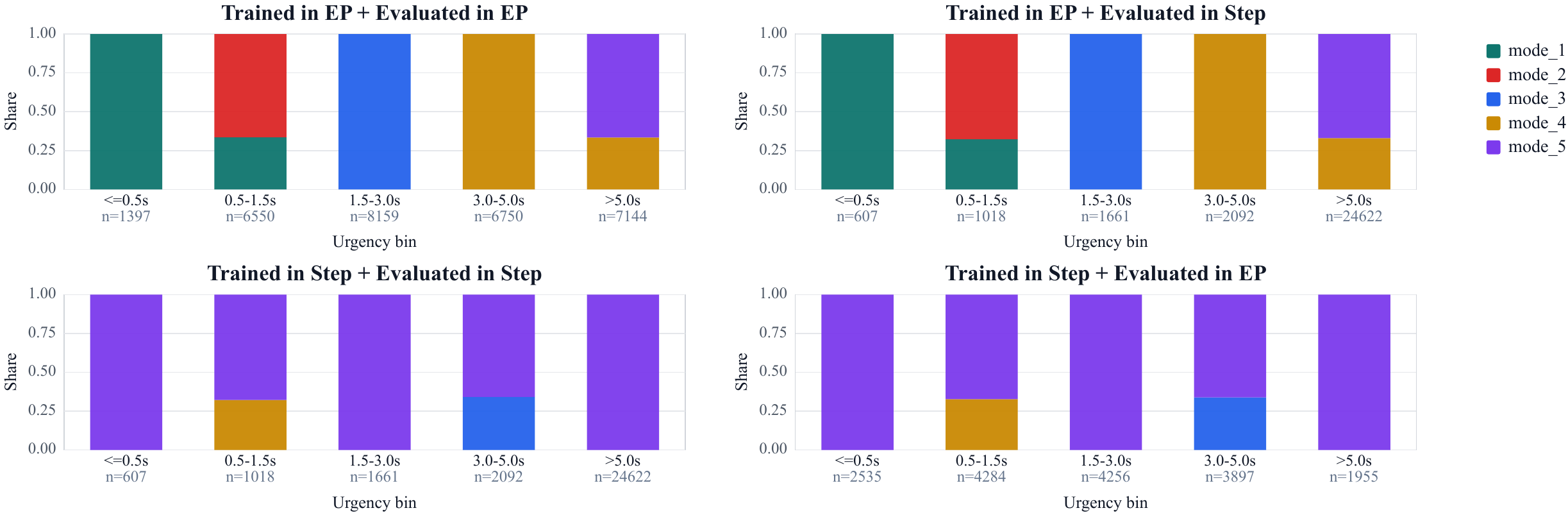}
\caption{Urgency-conditioned deliberation-mode distributions in the sequential-task setting. EP learns a monotone slack-aware schedule, while the step-trained policy remains biased toward the deepest mode even in tight-deadline states.}
\label{fig:toy1b-urgency-mode-distribution}
\end{figure}

\paragraph{Additional deliberation results.}
Tables~\ref{tab:toy1a-full} and~\ref{tab:toy1b-full} report the complete
cross-evaluation results for the deliberation experiments. The step-trained
policies perform best when deliberation is free, but their preference for slow
high-accuracy modes produces severe timeouts when evaluated in EP. The
urgency-conditioned mode distributions in
Figures~\ref{fig:toy1a-urgency-mode-distribution}
and~\ref{fig:toy1b-urgency-mode-distribution} show that EP learns to allocate
deeper deliberation primarily to states with sufficient slack. Cells report the
mean and half-width of a bootstrap $95\%$ confidence interval. The bootstrap is
computed over the $3{,}000$ evaluation episodes for the fixed training seed
$42$; therefore the intervals quantify evaluation-set variability for the
reported seed, not training-seed variability.

\begin{table}[ht]
\centering
\caption{Complete single-task urgency-aware deliberation results with bootstrap confidence intervals.}
\label{tab:toy1a-full}
\begin{adjustbox}{max width=\linewidth}
\begin{tabular}{lrrrr}
\toprule
\textbf{Training $\rightarrow$ evaluation} & \textbf{Mean return $\uparrow$} & \textbf{Success rate (\%) $\uparrow$} & \textbf{Timeout rate (\%) $\downarrow$} & \textbf{Mode 5 usage (\%)} \\
\midrule
EP $\rightarrow$ EP & \ci{0.80}{0.11} & \ci{46.7}{1.8} & \ci{0.0}{0.0} & \ci{21.2}{1.5} \\
EP $\rightarrow$ Step & \ci{0.77}{0.11} & \ci{46.1}{1.8} & \ci{0.0}{0.0} & \ci{19.3}{1.4} \\
Step $\rightarrow$ Step & \ci{2.54}{0.09} & \ci{75.6}{1.6} & \ci{0.0}{0.0} & \ci{93.5}{0.9} \\
Step $\rightarrow$ EP & \ci{-1.03}{0.08} & \ci{16.1}{1.3} & \ci{79.0}{1.5} & \ci{93.6}{0.9} \\
\bottomrule
\end{tabular}
\end{adjustbox}
\end{table}

\begin{table}[htbp]
\centering
\caption{Complete sequential urgency-aware deliberation results with bootstrap confidence intervals.}
\label{tab:toy1b-full}
\begin{adjustbox}{max width=\linewidth}
\begin{tabular}{lrrrr}
\toprule
\textbf{Training $\rightarrow$ evaluation} & \textbf{Mean return $\uparrow$} & \textbf{Success rate (\%) $\uparrow$} & \textbf{Timeout rate (\%) $\downarrow$} & \textbf{Mode 5 usage (\%)} \\
\midrule
EP $\rightarrow$ EP & \ci{10.07}{0.35} & \ci{50.1}{0.6} & \ci{3.7}{0.2} & \ci{15.8}{0.4} \\
EP $\rightarrow$ Step & \ci{21.54}{0.31} & \ci{69.2}{0.5} & \ci{0.0}{0.0} & \ci{55.0}{0.6} \\
Step $\rightarrow$ Step & \ci{25.63}{0.29} & \ci{76.0}{0.5} & \ci{0.0}{0.0} & \ci{96.5}{0.2} \\
Step $\rightarrow$ EP & \ci{-14.98}{0.24} & \ci{8.4}{0.4} & \ci{88.9}{0.5} & \ci{83.9}{0.5} \\
\bottomrule
\end{tabular}
\end{adjustbox}
\end{table}

The confidence intervals make the timing mismatch visible even under a fixed
training seed. In the single-task case, Step training achieves the highest
return when evaluated in the Step abstraction, but the same policy times out in
\ci{79.0}{1.5}\% of EP evaluations because it almost always selects the deepest
mode. EP training gives up some free-deliberation accuracy under Step
evaluation, but under EP evaluation it keeps mode-5 usage near
\ci{21.2}{1.5}\% and avoids timeout entirely in this run. The sequential setting
amplifies the same effect: Step $\rightarrow$ EP obtains \ci{-14.98}{0.24}
return with \ci{88.9}{0.5}\% timeouts, while EP $\rightarrow$ EP remains positive
at \ci{10.07}{0.35}. Thus, the result is not a small evaluation fluctuation:
when elapsed deliberation time is restored, the step-trained policy falls into
a consistently timeout-prone regime.

\paragraph{Patrol interruption experiments.}
Tables~\ref{tab:toy2a-config} and~\ref{tab:toy2b-config} specify the patrol
settings. The module-level experiment makes the checkpoint-handling duration and
alarm deadlines comparable in scale, so waiting for module completion can create
response latency. The state-level experiment decomposes handling into phases
with explicit interruption costs, making the decision depend on alarm urgency,
phase progress, distance, and sunk cost.

\paragraph{Hand-written interruption patches.}
The state-level experiment includes two non-learning interruption policies.
Patch is intentionally simple: it decides from the current phase and the
remaining alarm time, ignoring distance and phase progress. PatchPro is a
stronger hand-written policy that adds several checks before interrupting. It
first skips alarms that cannot be reached in time, responds immediately to very
urgent feasible alarms, responds during patrol navigation because no phase
progress is being abandoned, and otherwise checks whether the current handling
routine can be finished before still reaching the alarm. If finishing first is
not feasible, PatchPro uses phase-specific urgency thresholds, phase-progress
thresholds, and a cost-value check comparing the phase interruption cost with
the net value of resolving the alarm. In this experiment, the net alarm value is
the resolve reward plus the avoided expiration penalty, i.e., $25 - (-20)=45$.
Tables~\ref{tab:toy2b-patch-rules}
and~\ref{tab:toy2b-patchpro-thresholds} list these hand-written choices. These
thresholds are tied to the specific deadlines, rewards, interruption costs, and
phase durations used here; changing those parameters would require redesigning
or retuning the patch.

\begin{table}[htbp]
\caption{Decision rules used by the simple hand-written Patch baseline.}
\label{tab:toy2b-patch-rules}
\centering
\small
\begin{tabular}{@{}p{0.18\linewidth}p{0.28\linewidth}p{0.44\linewidth}@{}}
\hline
\textbf{Depth} & \textbf{Current state} & \textbf{Patch decision} \\
\hline
2 or 3 & No active alarm, navigating to alarm, or resolving alarm & Continue current action \\
2 or 3 & Patrol navigation & Respond to alarm \\
2 or 3 & Observe phase & Respond to alarm \\
2 & Commit phase & Respond only if remaining alarm time is at most $6$ ticks \\
3 & Verify phase & Respond only if remaining alarm time is at most $8$ ticks \\
3 & Commit phase & Continue checkpoint handling \\
\hline
\end{tabular}
\end{table}

\begin{table}[htbp]
\caption{Phase-specific thresholds used by PatchPro after feasibility, very-urgent, patrol-navigation, and finish-first checks. The very-urgent override responds whenever a feasible alarm has at most $3$ ticks remaining. The cost ratio requires the alarm's net value to exceed the phase interruption cost by the listed multiplier.}
\label{tab:toy2b-patchpro-thresholds}
\centering
\small
\begin{tabular}{@{}p{0.14\linewidth}p{0.22\linewidth}p{0.18\linewidth}p{0.20\linewidth}p{0.18\linewidth}@{}}
\hline
\textbf{Depth} & \textbf{Phase} & \textbf{Urgency threshold} & \textbf{Progress cutoff} & \textbf{Cost ratio} \\
\hline
2 & Observe & $16$ & $1.0$ & $2.0$ \\
2 & Commit & $8$ & $0.5$ & $3.5$ \\
3 & Observe & $18$ & $1.0$ & $2.0$ \\
3 & Verify & $12$ & $0.8$ & $3.0$ \\
3 & Commit & $8$ & $0.5$ & $4.5$ \\
\hline
\end{tabular}
\end{table}

\begin{table}[htbp]
\caption{Configuration for the module-level interruption experiment.}
\label{tab:toy2a-config}
\centering
\small
\begin{tabular}{@{}p{0.28\linewidth}p{0.64\linewidth}@{}}
\hline
\textbf{Parameter} & \textbf{Value} \\
\hline
Grid and checkpoints & $8\times 8$ grid; checkpoints at $(0,0)$, $(0,7)$, $(7,7)$, $(7,0)$ \\
Episode length & $1{,}000$ ticks \\
Checkpoint handling & $20$ ticks; checkpoint reward $+1$ \\
Alarm generation & Probability $0.15$ per tick; minimum distance $5$ \\
Alarm deadline & Uniformly sampled from $14$ to $22$ ticks \\
Alarm response & Navigation to alarm plus $2$ ticks to resolve \\
Alarm reward and penalties & $+25$ for resolve; $-20$ for expire; $-0.5$ per active tick \\
Decision interface & EP can interrupt at any tick; Loop decides only at module boundaries \\
Training and evaluation & $8{,}000$ training episodes; $3{,}000$ evaluation episodes; $\gamma=0.99$; seed $42$ \\
\hline
\end{tabular}
\end{table}

\begin{figure}[htbp]
\centering
\includegraphics[width=\linewidth]{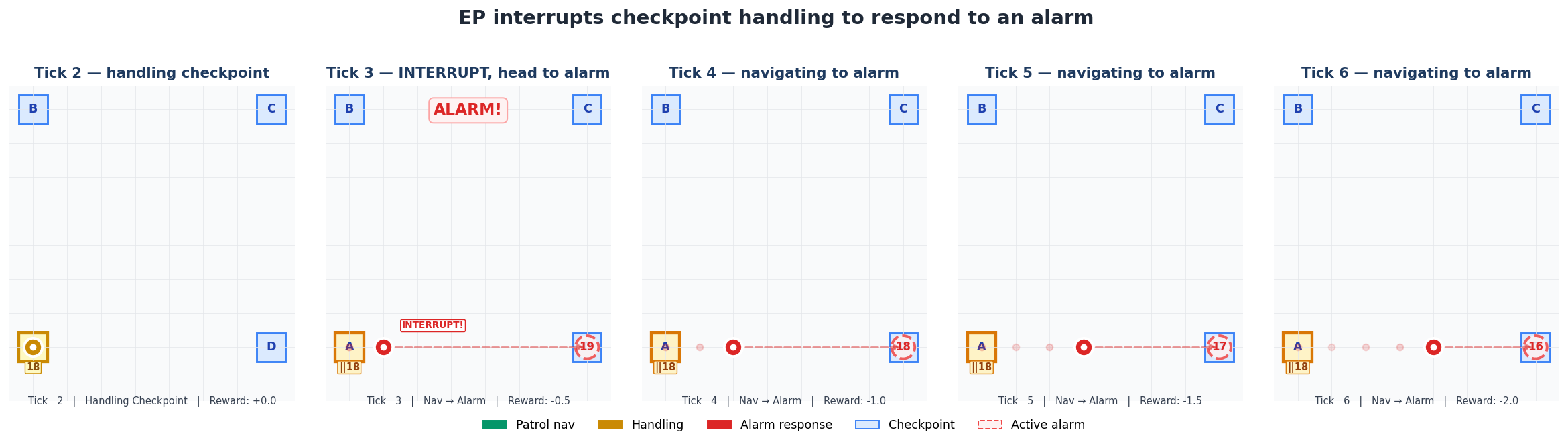}
\caption{EP interrupting an in-progress checkpoint handling. At tick $2$ the
agent is handling checkpoint A with $18$ ticks of work remaining and no active
alarm. At tick $3$ an alarm appears at checkpoint D with deadline $19$; because
EP can re-decide every tick, the upper level immediately switches the agent
from \textsc{handle\_ckpt} to \textsc{nav\_alarm} (the abandoned handling work
is preserved as \texttt{||18}). Ticks $4$--$6$ show the agent advancing one cell per
tick toward D while the alarm deadline counts down $18 \to 17 \to 16$, leaving
ample slack to resolve the alarm before it expires.}
\label{fig:toy2-ep-interrupt-trace}
\end{figure}

\begin{figure}[t]
\centering
\includegraphics[width=\linewidth]{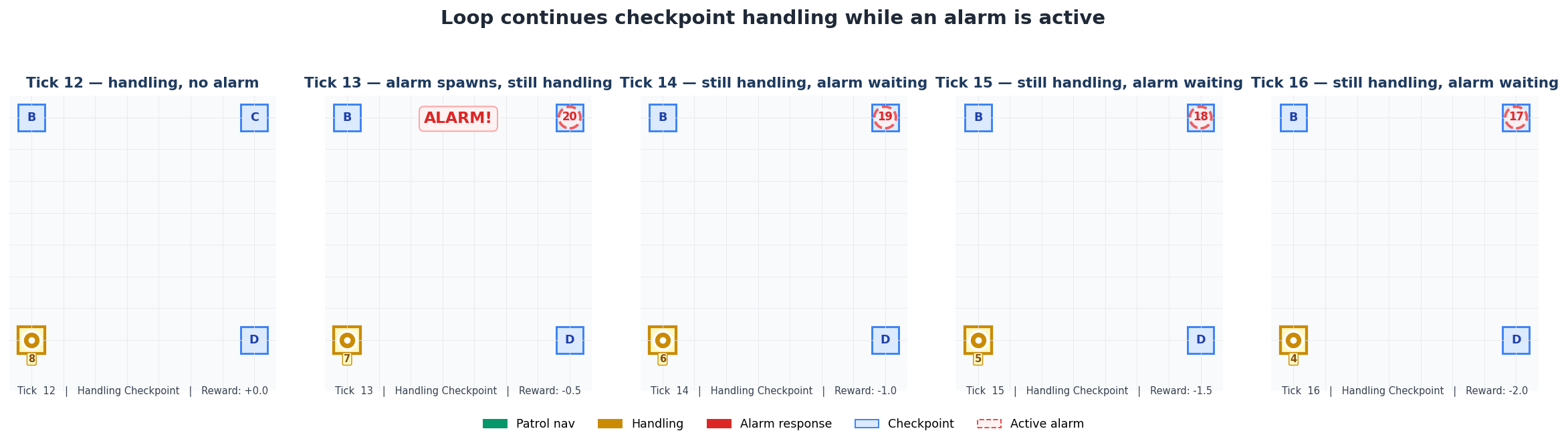}
\caption{Loop unable to interrupt an in-progress checkpoint handling. At tick
$12$ the agent is handling checkpoint A with no alarm; at tick $13$ an alarm
appears at C with deadline $20$. Because Loop only re-decides at module
boundaries, the agent stays in \textsc{handle\_ckpt} and the handling counter
keeps decrementing ($12 \to 11 \to 10 \to 9$) while the alarm deadline shrinks
in lockstep ($19 \to 18 \to 17$). The earliest tick at which Loop could have
responded would be after the remaining $9$ ticks of handling, by which point
the alarm has only $11$ ticks left and the agent must still cross the grid to
reach C.}
\label{fig:toy2-loop-stuck-trace}
\end{figure}

\paragraph{Module-level interruption traces.}
Figures~\ref{fig:toy2-ep-interrupt-trace} and~\ref{fig:toy2-loop-stuck-trace}
visualize five consecutive ticks from the module-level experiment with seed
$42$, drawn from the same evaluation distribution that produces
Table~\ref{tab:toy2a}. Each panel marks the agent's current module by color
(yellow for checkpoint handling, red for alarm response), shows the remaining
handling time as a small badge below the active checkpoint, and renders the
active alarm as a dashed circle annotated with its remaining deadline.

The two traces concretize the structural latency reflected in
Table~\ref{tab:toy2a}: EP commits to alarm response at the first available
tick, keeping its mean ticks-per-alarm at $10.9$ and its expire rate at
$8.0\%$, whereas Loop can only act at module boundaries, so any alarm that
spawns mid-handling is forced to wait out the remaining handling duration,
inflating ticks per alarm to $14.4$ and expiring nearly half of all alarms
($47.1\%$). For each method, the displayed sequence is the first event in the
evaluation episode that meets the corresponding criterion: an interruption of
checkpoint handling for EP, and an alarm spawn during checkpoint handling for
Loop.

\begin{table}[htbp]
\caption{Configuration for the state-level interruption experiment.}
\label{tab:toy2b-config}
\centering
\small
\begin{tabular}{@{}p{0.28\linewidth}p{0.64\linewidth}@{}}
\hline
\textbf{Parameter} & \textbf{Value} \\
\hline
Grid and checkpoints & $8\times 8$ grid; checkpoints at $(1,1)$, $(1,6)$, $(6,6)$, $(6,1)$ \\
Episode length & $1{,}000$ ticks \\
Depth-2 phases & Observe: $4$ ticks, cost $-1$; Commit: $10$ ticks, cost $-5$ \\
Depth-3 phases & Observe: $3$ ticks, cost $-1$; Verify: $5$ ticks, cost $-4$; Commit: $10$ ticks, cost $-7$ \\
Alarm generation & Probability $0.07$ per tick; minimum distance $5$ \\
Alarm deadline & Uniformly sampled from $7$ to $12$ ticks \\
Alarm response & Navigation to alarm plus $2$ ticks to resolve \\
Alarm reward and penalties & $+25$ for resolve; $-20$ for expire; $-0.5$ per active tick \\
Decision methods & EP, Loop, Patch, and PatchPro \\
Training and evaluation & $8{,}000$ training episodes; $3{,}000$ evaluation episodes; $\gamma=0.99$; seed $42$ \\
\hline
\end{tabular}
\end{table}

\paragraph{Additional patrol results.}
Tables~\ref{tab:toy2a-ci}, \ref{tab:toy2b-depth2-ci}, and
\ref{tab:toy2b-depth3-ci} report bootstrap confidence intervals for the patrol
experiments. As above, the bootstrap resamples the $3{,}000$ evaluation
episodes generated from the fixed training seed $42$. For rates in the patrol
experiments, each bootstrap sample recomputes the ratio from aggregate counts,
e.g., resolved alarms divided by total alarms, rather than averaging per-episode
ratios.

\begin{table}[htbp]
\centering
\caption{Module-level patrol interruption results. Cells report mean $\pm$ half-width of the bootstrap 95\% confidence interval.}
\label{tab:toy2a-ci}
\begin{adjustbox}{max width=\linewidth}
\begin{tabular}{lrrrr}
\toprule
\textbf{Method} & \textbf{Mean return $\uparrow$} & \textbf{Resolve rate (\%) $\uparrow$} & \textbf{Expire rate (\%) $\downarrow$} & \textbf{Ticks per alarm $\downarrow$} \\
\midrule
EP & \ci{896.59}{4.27} & \ci{90.9}{0.1} & \ci{8.0}{0.1} & \ci{10.94}{0.02} \\
Loop & \ci{-163.55}{5.53} & \ci{51.5}{0.2} & \ci{47.1}{0.2} & \ci{14.39}{0.02} \\
\bottomrule
\end{tabular}
\end{adjustbox}
\end{table}

\begin{table}[ht]
\centering
\caption{State-level patrol interruption results with two handling phases. Cells report mean $\pm$ half-width of the bootstrap 95\% confidence interval.}
\label{tab:toy2b-depth2-ci}
\begin{adjustbox}{max width=\linewidth}
\begin{tabular}{lrrrr}
\toprule
\textbf{Method} & \textbf{Mean return $\uparrow$} & \textbf{Resolve rate (\%) $\uparrow$} & \textbf{Expire rate (\%) $\downarrow$} & \textbf{Interrupt cost} \\
\midrule
EP & \ci{175.09}{5.27} & \ci{62.8}{0.3} & \ci{36.5}{0.3} & \ci{61.70}{0.50} \\
PatchPro & \ci{25.06}{5.47} & \ci{53.9}{0.3} & \ci{45.4}{0.3} & \ci{27.99}{0.36} \\
Patch & \ci{-156.54}{5.65} & \ci{47.6}{0.3} & \ci{51.6}{0.3} & \ci{63.73}{0.49} \\
Loop & \ci{-344.96}{5.38} & \ci{34.9}{0.3} & \ci{64.4}{0.3} & \ci{0.00}{0.00} \\
\bottomrule
\end{tabular}
\end{adjustbox}
\end{table}

\begin{table}[htbp]
\centering
\caption{State-level patrol interruption results with three handling phases. Cells report mean $\pm$ half-width of the bootstrap 95\% confidence interval.}
\label{tab:toy2b-depth3-ci}
\begin{adjustbox}{max width=\linewidth}
\begin{tabular}{lrrrr}
\toprule
\textbf{Method} & \textbf{Mean return $\uparrow$} & \textbf{Resolve rate (\%) $\uparrow$} & \textbf{Expire rate (\%) $\downarrow$} & \textbf{Interrupt cost} \\
\midrule
EP & \ci{143.31}{5.03} & \ci{62.4}{0.2} & \ci{36.8}{0.3} & \ci{80.62}{0.63} \\
PatchPro & \ci{68.84}{5.50} & \ci{57.2}{0.3} & \ci{42.0}{0.3} & \ci{48.44}{0.53} \\
Patch & \ci{-101.05}{5.27} & \ci{48.4}{0.3} & \ci{50.8}{0.3} & \ci{27.10}{0.32} \\
Loop & \ci{-298.66}{5.36} & \ci{37.5}{0.3} & \ci{61.7}{0.3} & \ci{0.00}{0.00} \\
\bottomrule
\end{tabular}
\end{adjustbox}
\end{table}

The module-level results show a large, statistically stable separation between
EP and Loop under the fixed seed-42 policy. EP resolves \ci{90.9}{0.1}\% of
alarms and keeps the expiration rate to \ci{8.0}{0.1}\%, while Loop resolves
only \ci{51.5}{0.2}\% and expires \ci{47.1}{0.2}\%. The narrow intervals arise
because each evaluation contains many alarm events; the performance gap is much
larger than the evaluation uncertainty. This supports the qualitative trace in
Figures~\ref{fig:toy2-ep-interrupt-trace} and~\ref{fig:toy2-loop-stuck-trace}:
the main loss for Loop is not a weaker low-level primitive, but delayed access
to the high-level decision point.

The state-level experiments add interruption costs and finer handling phases,
but the same ordering remains. At depth 2, EP exceeds PatchPro by about $150$
return points and resolves about $8.9$ percentage points more alarms, despite
paying higher interruption cost. At depth 3, PatchPro improves relative to the
simple Patch by using hand-designed reachability and progress checks, but EP
still has the best return and the lowest expiration rate. Loop has zero
interruption cost by construction, yet this is dominated by missed alarms:
waiting until phase boundaries produces the lowest return and highest expiration
rate in both depth settings. Overall, the CI tables indicate that EP's advantage
comes from exposing interruption as part of the interaction process, not from
noise in a particular evaluation batch.

% \FloatBarrier
\section{Digital Assistant Experiment Configuration Details}
\label{app:digital-assistant-config}

This appendix gives the numerical configuration used for the digital assistant
with asynchronous emails experiment in Section~\ref{sec:digital_assistant_async_emails}.
The experiment evaluates whether EP's event-driven interface improves behavior
when observations arrive while an LLM-generated main-task action is still
unfolding. All compared methods share the same task distribution, email stream,
action primitives, action durations, and scoring rule. They differ only in the
decision interface used to observe and react to asynchronous email events.

\paragraph{Episode and environment.}
Each episode lasts 90 simulated time units. At the start of an episode, the
assistant receives one workplace-style main task. Main-task text generation is
replayed from a cached output database, and replay consumes simulated time at a
rate of 0.05 time units per generated token. During the episode, emails arrive
according to a stochastic arrival process with rate 0.2. Each email has an
urgency level, a deadline, and an action recommendation distribution used by
the triage policy. Urgency levels are sampled with probabilities 0.4, 0.4, and
0.2 for high-, medium-, and low-urgency emails. Deadline slack is sampled from
the ranges shown in Table~\ref{tab:digital-assistant-env-config}.

\begin{table}[t]
\caption{Environment configuration for the digital assistant with asynchronous emails experiment.}
\label{tab:digital-assistant-env-config}
\centering
\small
\begin{tabular}{@{}p{0.30\linewidth}p{0.62\linewidth}@{}}
\hline
\textbf{Parameter} & \textbf{Value} \\
\hline
Episode horizon & 90 simulated time units \\
Email arrival rate & 0.2 \\
Token time cost & 0.05 simulated time units per token \\
Main-task target steps & 4 \\
Main-task selection & Seeded hash over cached task outputs \\
Email scenario & Default cached email database \\
Urgency probabilities & High 0.4, medium 0.4, low 0.2 \\
High-urgency deadline slack & Uniform from 5 to 15 time units \\
Medium-urgency deadline slack & Uniform from 15 to 25 time units \\
Low-urgency deadline slack & Uniform from 25 to 35 time units \\
Check inbox duration & 0.1 \\
Open message duration & 0.1 \\
Handle email duration & 1.0 \\
Set reminder duration & 0.1 \\
Return to main task duration & 0.1 \\
Triage stream duration & 0.2 \\
\hline
\end{tabular}
\end{table}

\paragraph{Compared decision interfaces.}
The experiment compares EP, Step, and Patch. Step is a synchronized loop that
does not observe new email arrivals while a main-task unit is being generated;
it checks the inbox only after the current main-task unit finishes. Patch adds
a fixed polling interval of 15 simulated time units during main-task generation.
At each polling point it interrupts, checks the active emails, processes them,
and then resumes the main task. EP treats main-task generation as an
interruptible timed action: email arrivals and reminder events can occur while
the main-task stream is unfolding, and the agent can interrupt the stream,
triage the email, act on it, and later resume the partially completed main-task
output.

\paragraph{Run settings.}
We evaluate four settings, crossing two email-triage model settings with two
main-task decompositions. The milestone setting uses the milestone-level
subtasks in the task definition. The single setting collapses the main task
into one longer unit. Each setting uses 200 seeds and 5 episodes per seed for
each method, giving 1000 episodes per method and 3000 episodes per setting.

\paragraph{Scoring rule.}
Utility combines main-task progress,
main-task completion, timely email handling, timeout penalties, and small
switching/interruption penalties:
\begin{align}
U
&=
0.4\,(\text{main progress units})
+0.8\,\mathbb{1}\{\text{main completed}\}
+0.4\,(\text{emails handled before deadline})
\nonumber\\
&\quad
-0.5\,(\text{timed-out emails})
-0.005\,(\text{focus switches})
-0.005\,(\text{main-stream interruptions}).
\end{align}
\paragraph{Main and email score definitions.}
The balanced score uses two normalized components that are separate from the
additive utility above. The main-task component is the fraction of the target
main-task progress completed:
\begin{equation}
    \text{Main}
    =
    \min\!\left(1,\,
    \frac{\text{main progress units}}{\text{main target progress units}}
    \right).
\end{equation}
In these runs the target progress is 4 units. The main completion bonus in the
utility is not added to this normalized score; it only contributes to $U$.

The email component is an urgency-weighted on-time completion rate. Let
$\mathcal{E}$ be the set of emails that arrive in the episode, let $w(e)$ be
the urgency weight of email $e$, and let $\mathbb{1}_{\mathrm{on\text{-}time}}(e)$
indicate that the email was handled no later than its deadline. Then
\begin{equation}
    \text{Email}
    =
    \frac{
        \sum_{e\in\mathcal{E}} w(e)\,
        \mathbb{1}_{\mathrm{on\text{-}time}}(e)
    }{
        \sum_{e\in\mathcal{E}} w(e)
    }.
\end{equation}
Emails handled after their deadline and emails that time out both contribute 0
to the numerator. The urgency weights are 1.2, 0.5, and 0.1 for high-, medium-,
and low-urgency emails. If no email arrives, the email component is defined as
1.0.

The final balanced score combines the normalized main-task score and the
urgency-weighted email completion score using complementary weights
$\alpha_{\mathrm{main}}$ and $1-\alpha_{\mathrm{main}}$ before applying an
additional imbalance penalty:
\begin{equation}
    \text{Balanced}
    =
    \alpha_{\mathrm{main}}\, \text{Main}
    +
    (1-\alpha_{\mathrm{main}})\, \text{Email}
    -
    \lambda\, |\text{Main}-\text{Email}|.
\end{equation}
In the reported configuration, $\alpha_{\mathrm{main}}=0.6$,
$1-\alpha_{\mathrm{main}}=0.4$, and $\lambda=0.5$.

\paragraph{Qwen3.5-Plus results.}
Table~\ref{tab:digital_assistant_async_email_qwen} reports the Qwen3.5-Plus results omitted from the main text for space. The ordering matches the DeepSeek-V3.2 results: EP performs best in both decompositions, Periodic Poll improves over AgentLoop but remains below EP, and AgentLoop preserves main-task completion at the cost of email responsiveness.

\begin{table}[t]
\centering
\small
\caption{Digital assistant with asynchronous emails. Utility, Balanced, Timeout, Latency, and Main are reported as mean $\pm$ half-width of bootstrap 95\% CI (1000 resamples, seed 42). Utility, Balanced, and Main are higher-is-better; Timeout and Latency are lower-is-better. Each run contains 1000 episodes per method.}
\label{tab:digital_assistant_async_email_qwen}
\begin{adjustbox}{max width=\linewidth}
\begin{tabular}{llrrrrr}
\toprule
\textbf{Setting} & \textbf{Method} &
\textbf{Utility} &
\textbf{Balanced} &
\textbf{Timeout} &
\textbf{Latency} &
\textbf{Main} \\
\midrule
DeepSeek-V3.2-single & EP    & \ci{3.023}{0.102} & \ci{0.792}{0.010} & \ci{0.315}{0.010} & \ci{6.92}{0.20} & \ci{0.984}{0.008} \\
DeepSeek-V3.2-single & Periodic Poll & \ci{2.016}{0.119} & \ci{0.619}{0.013} & \ci{0.426}{0.013} & \ci{10.15}{0.17} & \ci{0.983}{0.008} \\
DeepSeek-V3.2-single & AgentLoop  & \ci{0.730}{0.117} & \ci{0.382}{0.014} & \ci{0.598}{0.015} & \ci{13.58}{0.27} & \ci{1.000}{0.000} \\
\midrule
Qwen3.5-Plus-single & EP    & \ci{3.643}{0.087} & \ci{0.834}{0.009} & \ci{0.263}{0.010} & \ci{7.28}{0.20} & \ci{0.988}{0.007} \\
Qwen3.5-Plus-single & Periodic Poll & \ci{2.563}{0.110} & \ci{0.660}{0.013} & \ci{0.375}{0.012} & \ci{10.47}{0.17} & \ci{0.981}{0.009} \\
Qwen3.5-Plus-single & AgentLoop  & \ci{0.776}{0.118} & \ci{0.383}{0.014} & \ci{0.593}{0.015} & \ci{13.70}{0.28} & \ci{1.000}{0.000} \\
\midrule
DeepSeek-V3.2-milestones & EP    & \ci{3.113}{0.105} & \ci{0.818}{0.009} & \ci{0.305}{0.011} & \ci{6.91}{0.19} & \ci{0.995}{0.002} \\
DeepSeek-V3.2-milestones & Periodic Poll & \ci{1.775}{0.115} & \ci{0.606}{0.014} & \ci{0.444}{0.014} & \ci{11.33}{0.21} & \ci{0.998}{0.002} \\
DeepSeek-V3.2-milestones & AgentLoop  & \ci{1.591}{0.123} & \ci{0.569}{0.014} & \ci{0.467}{0.015} & \ci{11.83}{0.23} & \ci{1.000}{0.000} \\
\midrule
Qwen3.5-Plus-milestones & EP    & \ci{3.755}{0.089} & \ci{0.859}{0.007} & \ci{0.253}{0.010} & \ci{7.47}{0.19} & \ci{0.995}{0.002} \\
Qwen3.5-Plus-milestones & Periodic Poll & \ci{2.029}{0.108} & \ci{0.632}{0.013} & \ci{0.417}{0.014} & \ci{11.62}{0.21} & \ci{0.998}{0.001} \\
Qwen3.5-Plus-milestones & AgentLoop  & \ci{1.767}{0.118} & \ci{0.590}{0.014} & \ci{0.446}{0.015} & \ci{12.15}{0.22} & \ci{1.000}{0.000} \\
\bottomrule
\end{tabular}
\end{adjustbox}
\end{table}

Averaged over the two Qwen3.5-Plus settings, EP reduces timeout rate from 0.520 for AgentLoop to 0.258 and reduces first-response latency from 12.93 to 7.38. Periodic Poll reaches 0.396 timeout rate and 11.05 latency. EP improves utility over AgentLoop by 2.867 points in \emph{single} and 1.988 points in \emph{milestones}; the corresponding balanced-score gains are 0.451 and 0.269. These results show that the benefit of event-time email observation is not specific to the DeepSeek-V3.2 triage model.

\begin{table}[t]
\caption{Scoring configuration used for the reported digital assistant results.}
\label{tab:digital-assistant-scoring-config}
\centering
\small
\begin{tabular}{@{}p{0.34\linewidth}p{0.54\linewidth}@{}}
\hline
\textbf{Scoring parameter} & \textbf{Value} \\
\hline
Main progress reward & 0.4 per progress unit \\
Main completion bonus & 0.8 \\
On-time email handling reward & 0.4 per email \\
Late email handling reward & 0.0 \\
Timeout penalty & 0.5 per timed-out email \\
Focus switch penalty & 0.005 per switch \\
Main-stream interruption penalty & 0.005 per interruption \\
High-, medium-, low-urgency weights & 1.2, 0.5, 0.1 \\
Balanced main weight & 0.6 \\
Balanced gap penalty & 0.5 \\
\hline
\end{tabular}
\end{table}

\FloatBarrier
\section{Embodied Scheduling Task Details and Additional Results}
\label{app:embodied-scheduling}

This appendix gives implementation details for the embodied scheduling experiments in
Section~\ref{sec:embodied_scheduling}. The benchmark couples a kitchen process with
an online tutoring process on a single simulated clock. All compared interfaces use the
same dishes, tutoring problems, action vocabulary, random seeds, time model, and scoring
rule. They differ only in the temporal interface exposed to the controller: when cooking
signals become visible during tutoring, whether tutoring may be interrupted at the signal
time, and whether the immediate kitchen reaction is chosen by the controller or by an
engine-side reflex.

\paragraph{Shared task model and utility.}
Each episode places an LLM controller in a kitchen-plus-tutoring simulator. The controller
must cook dishes and make progress on tutoring problems along the same timeline. Cooking
actions consume simulated time and may create future cooking signals. Tutoring also consumes
time, with one tick corresponding to one generated tutor token. The controller's own completion
tokens also consume simulated ticks. This shared clock makes kitchen timing, tutoring progress,
and controller deliberation directly comparable.

All reported embodied results use
\[
U
=
\sum_{d \in \mathcal{D}} 5 q_d
+
5 N_{\mathrm{tutor}}
-
0.003 T
-
0.75 N_{\mathrm{switch}}
-
0.05 C_{\mathrm{check}},
\]
where \(q_d\) is the quality of dish \(d\), \(N_{\mathrm{tutor}}\) is the number of completed
tutor problems, \(T\) is total elapsed time in ticks, \(N_{\mathrm{switch}}\) is the number of
kitchen--tutor mode switches, and \(C_{\mathrm{check}}\) is the number of explicit check actions.
Critical dishes receive \(q_d=1\) if they are finished within their ideal response window and
\(q_d=0\) otherwise. For the reported one-dish and multi-dish runs, soft dishes use flat
keep-warm scoring: a finish between signal onset and the keep-warm deadline receives full
quality \(q_d=1\), while a finish after the keep-warm deadline receives \(q_d=0\). No separate
judge model is used in these experiments, so tutoring contributes to utility through completed
tutor problems rather than answer correctness.

\paragraph{Shared controller prompt template.}
The one-dish and multi-dish settings use the same controller prompt scaffold. The system prompt first
establishes the controller's role as an assistant embedded in a kitchen-plus-tutoring simulator,
operating over a single shared timeline and a single shared interaction history. It then states the
clock semantics: one tick corresponds to one generated tutor token, and the controller's own completion
tokens also consume simulated ticks. The prompt next defines the legal XML action protocol, including
kitchen actions for preparing, staging, and finishing dishes; tutoring actions; waiting; and terminating
the episode.

The prompt also specifies the semantics of tutoring segments. A \texttt{\textless tutor/\textgreater}
action without an explicit \texttt{tokens} attribute defaults to the remaining length of the current
tutor problem, unless the interface returns control earlier. A \texttt{tokens=N} attribute is interpreted
as a voluntary upper bound unless the variant semantics interrupt the segment sooner. After these
interface-independent rules, the prompt presents the visible task state, including the currently
available kitchen and tutoring tasks, followed by utility guidance describing dish rewards, tutor
completion rewards, elapsed-time cost, and kitchen--tutor switch cost.

Finally, the prompt reminds the controller that tutoring is online: when a tutoring segment is
interrupted and later resumed, the model should continue the unfinished idea rather than restart the
explanation. The prompt ends with a variant-specific rule describing when cooking signals become
visible, whether tutoring may be interrupted at the signal time, and whether the immediate kitchen
reaction is selected by the controller or by an engine-side reflex.

The abstract shared controller template is:

\begin{promptbox}
You are an assistant embedded in a kitchen + tutoring simulator.
You operate on one shared timeline and one shared interaction history.
1 tick = 1 tutor token. Your own completion tokens also consume ticks.

[Action protocol]
<action kind="start_prep" dish="X"/>
<action kind="put_on_stove" dish="X"/>
<action kind="finish" dish="X"/>
<tutor .../>
<wait/>
<done/>

[Tutor segment semantics]
- <tutor/> without tokens defaults to the remaining part of the
  targeted problem, unless the interface returns control earlier.
- tokens=N is a voluntary upper bound unless the variant semantics
  override it.

[Visible task state]
- dish/task descriptors
- tutor problem descriptors

[Utility guidance]
- finish critical dish in-window: +5
- finish soft dish: +5 * quality
- complete tutor problem: +5
- world-clock cost: -0.003 per tick
- switch cost: -0.75 per kitchen<->tutor switch

[Online tutoring reminder]
- if tutoring is interrupted and later resumed, continue the
  unfinished idea instead of restarting

[Variant rule]
- interface-specific visibility, interruption, and reaction rule
\end{promptbox}

The one-dish and multi-dish prompts instantiate only the visible-task-state block differently.
In the one-dish setting, the prompt contains exactly one dish record and one tutor-problem record:
\begin{promptbox}
Dish:
- name=...
- criticality=...
- prep=...
- signal=..., mean=..., std=...
- ideal_window_width=...
- finish=...
- keep_warm=...        [soft dishes only]

Tutor problem:
- q="Q1"
- title=...
- total_tokens=...
\end{promptbox}
For soft one-dish cases, the visible state also includes a soft-dish note explaining that the dish
keeps full quality until its keep-warm deadline.

In the multi-dish setting, the same block expands the dish and tutor fields into lists and adds
scheduling guidance:
\begin{promptbox}
Dishes:
- C1 (...) [critical] prep=..., signal=..., window=..., finish=...
- S1 (...) [soft] prep=..., signal=..., window=..., finish=..., keep_warm=...
- ...

Tutor tasks:
- q="Q1" total_tokens=... title=...
- q="Q2" total_tokens=... title=...
- ...

Additional scheduling guidance:
- compare ticks_to_window_end and ticks_to_keep_warm_end
- reason about stove contention
- do not tunnel on late dishes while earlier-signal dishes remain unstaged
- selective reprioritization matters more than raw reflex speed
\end{promptbox}

\paragraph{Online tutor-generation subsystem.}
The controller prompt is separate from the prompt used to generate online tutoring text.
When online tutoring is enabled, the tutor-generation prompt uses the same structure in both
the one-dish and multi-dish settings: (1) a system message instructing the model to output
tutoring prose only and not mention the simulator, kitchen events, XML tags, or interruption
mechanics; (2) a user message containing the problem label, title, task brief, target length
budget, a status flag indicating whether this is a fresh explanation or a continuation,
task-specific guidance, the tail of the prior transcript, and a clipped summary of the recent
shared context; and (3) a final instruction to continue the tutoring text while staying on the
academic task even if recent context mentions kitchen events.

The abstract online tutor template is:

\begin{promptbox}
System:
You are generating live tutoring text inside a kitchen-plus-tutoring
benchmark.
Output tutoring prose only.
Do not mention the simulator, the kitchen, XML tags, or interruption
mechanics.
If prior transcript exists, continue the unfinished idea naturally
instead of restarting.

User:
Problem label: ...
Title: ...
Task brief: ...
Target length: about ... characters.
Task status: [new task / continue previous task]
Guidance: ...
Prior transcript tail:
...
Recent shared context:
...
Continue the tutoring text now. Stay on the academic tutoring task
even if the recent context mentions kitchen events.
\end{promptbox}

This separation ensures that interruptions affect scheduling and state visibility without turning
the tutor text itself into commentary about the simulator.

\paragraph{Compared interaction interfaces.}
The four interfaces differ only in the final variant-specific rule appended to the shared controller
prompt.

\textbf{AgentLoop} is the synchronized tutoring-step baseline. Cooking signals are not surfaced
while the controller is inside a tutoring segment. If the controller emits
\texttt{\textless tutor/\textgreater} without a token bound, tutoring runs until the current tutor
problem finishes or the segment otherwise returns. Any cooking signal that fires during that interval
becomes visible only after the tutoring segment boundary. The controller then chooses the next
kitchen or tutoring action from the delayed visible state.

\textbf{Periodic Poll} is the polling-based baseline. The controller may still issue a long tutor
request, but the runtime checks for newly fired cooking signals at fixed polling boundaries. In the
reported runs, the polling interval is 40 ticks. If a signal is detected at a polling boundary, control
returns to the controller with the updated visible state. The controller, not the engine, decides
whether to finish a dish, defer a soft dish, resume tutoring, wait, or take another admissible action.
This interface therefore reduces boundary-only delay, but signal visibility is still quantized by the
polling interval.

\textbf{ReflexPatch} is an engine-assisted reference. A cooking signal can interrupt tutoring
immediately, but before the next controller decision the engine automatically applies a default
highest-priority kitchen reaction. The controller then observes the post-reflex state and continues.
This baseline measures the benefit of immediate reflexive handling, but it is not a standard
agent-controlled policy because the crucial immediate reaction is hard-coded by the runtime.

\textbf{EP} is the event-driven controller interface. Newly fired cooking signals are surfaced at
event time during tutoring, and control returns to the controller at the interruption point. The signal
enters the same decision context as other observations, and the controller decides whether to finish
a dish immediately, defer a soft event, resume tutoring, wait, or take another admissible action.
EP therefore tests whether immediate event visibility plus agent-side control can recover the
benefits of timely reaction without replacing the controller's scheduling decision by an engine-side
reflex.

\paragraph{One-dish setting.}
The one-dish-one-tutor setting contains exactly one dish, one tutor problem, and stove capacity
\(S_{\mathrm{stove}}=1\). It isolates response latency by removing most multi-process scheduling
ambiguity. We use four randomized dish families:
\texttt{critical\_narrow}, \texttt{critical\_medium}, \texttt{critical\_wide}, and
\texttt{soft\_keepwarm}. Critical-narrow dishes use prep time \(10\)--\(16\), signal offset
\(140\)--\(240\), finish time \(6\)--\(9\), and ideal window width \(20\)--\(28\). Critical-medium
dishes use prep time \(12\)--\(18\), signal offset \(190\)--\(320\), finish time \(7\)--\(10\), and
ideal window width \(30\)--\(48\). Critical-wide dishes use prep time \(16\)--\(24\), signal offset
\(280\)--\(520\), finish time \(8\)--\(12\), and ideal window width \(60\)--\(100\). Soft keep-warm
dishes use prep time \(12\)--\(18\), signal offset \(140\)--\(280\), finish time \(12\)--\(18\),
ideal window width \(16\)--\(28\), and keep-warm slack \(220\)--\(380\).

Tutor problems are drawn from four length families: \texttt{xs} with \(80\)--\(100\) tokens,
\texttt{short} with \(120\)--\(200\) tokens, \texttt{medium} with \(220\)--\(380\) tokens, and
\texttt{long} with \(520\)--\(900\) tokens. Prompts are sampled from a generic mathematics
tutoring pool. The reported one-dish run uses
\(4\) dish families \(\times\) \(4\) tutor families \(\times\) \(2\) randomized structures per cell
\(\times\) \(4\) seeds \((7,11,17,23)\), for a total of \(128\) cases. The runtime configuration
uses \(\texttt{polling\_interval}=40\), no default tutor-token budget, no tutor-segment cap,
\(\texttt{max\_controller\_calls}=40\), and a time budget of \(8000\) ticks. To avoid a fully
deterministic signal boundary on synthetic whistle dishes, whistle-based signal offsets receive
Gaussian jitter with standard deviation \(3.0\).

% Existing Table 16 goes here.
\begin{table}[ht]
\centering
\caption{One-dish-one-tutor embodied scheduling results. Cells report mean $\pm$ half-width of the bootstrap 95\% confidence interval.}
\label{tab:embodied-one-dish-aggregate}
\begin{adjustbox}{max width=\linewidth}
\begin{tabular}{lrrrrrr}
\toprule
\textbf{Method} & \textbf{$n$} & \textbf{Utility $\uparrow$} & \textbf{Dish done $\uparrow$} & \textbf{Vis. delay $\downarrow$} & \textbf{Finish delay $\downarrow$} & \textbf{Switches} \\
\midrule
AgentLoop & 128 & \ci{3.604}{0.414} & \ci{0.266}{0.075} & \ci{224.849}{50.247} & \ci{227.750}{49.467} & \ci{2.000}{0.000} \\
Periodic Poll & 128 & \ci{5.598}{0.335} & \ci{0.922}{0.043} & \ci{4.659}{1.540} & \ci{17.151}{1.596} & \ci{3.273}{0.192} \\
ReflexPatch & 128 & \ci{5.254}{0.306} & \ci{0.938}{0.043} & \ci{4.468}{3.625} & \ci{15.960}{3.137} & \ci{3.891}{0.176} \\
EP & 128 & \ci{5.731}{0.313} & \ci{0.938}{0.043} & \ci{3.057}{2.789} & \ci{15.484}{3.000} & \ci{3.352}{0.207} \\
\bottomrule
\end{tabular}
\end{adjustbox}
\end{table}
% Suggested label: \label{tab:embodied-one-dish-aggregate}

The aggregate one-dish results in Table~\ref{tab:embodied-one-dish-aggregate} show three patterns.
First, AgentLoop is consistently weak because its signal-to-visibility delay is dominated by tutor
segment boundaries, which directly lowers dish completion. Second, Periodic Polling recovers most
of the benefit of interruptibility in this low-complexity setting: with only one dish and no burner
contention, polling every 40 ticks is often sufficient to surface the relevant signal before the dish
is irrecoverable. Third, EP obtains the best overall one-dish utility and the lowest average visibility
and finish delays, but the margin over Periodic Polling and ReflexPatch should be interpreted as
modest. The stronger conclusion is that boundary-only visibility fails, while both event-time
visibility and sufficiently frequent polling substantially improve embodied timing in the one-dish
regime.

% Existing Table 17 goes here.
\begin{table}[ht]
\centering
\caption{One-dish results by dish family. Cells report mean $\pm$ half-width of the bootstrap 95\% confidence interval.}
\label{tab:embodied-one-dish-family}
\begin{adjustbox}{max width=\linewidth}
\begin{tabular}{llrrrr}
\toprule
\textbf{Family} & \textbf{Method} & \textbf{$n$} & \textbf{Utility $\uparrow$} & \textbf{Vis. delay $\downarrow$} & \textbf{Finish delay $\downarrow$} \\
\midrule
critical\_medium & AgentLoop & 32 & \ci{3.463}{0.812} & \ci{182.105}{93.947} & \ci{194.105}{92.448} \\
critical\_medium & Periodic Poll & 32 & \ci{5.871}{0.641} & \ci{3.161}{2.903} & \ci{15.161}{2.952} \\
critical\_medium & ReflexPatch & 32 & \ci{5.203}{0.594} & \ci{2.759}{4.138} & \ci{14.759}{4.138} \\
critical\_medium & EP & 32 & \ci{6.189}{0.638} & \ci{2.500}{3.750} & \ci{14.500}{3.750} \\
\midrule
critical\_narrow & AgentLoop & 32 & \ci{3.731}{0.877} & \ci{232.286}{88.738} & \ci{214.542}{83.980} \\
critical\_narrow & Periodic Poll & 32 & \ci{5.123}{1.023} & \ci{5.844}{3.516} & \ci{17.844}{3.484} \\
critical\_narrow & ReflexPatch & 32 & \ci{5.447}{0.383} & \ci{0.000}{0.000} & \ci{12.000}{0.000} \\
critical\_narrow & EP & 32 & \ci{6.306}{0.595} & \ci{3.161}{4.742} & \ci{14.750}{5.063} \\
\midrule
critical\_wide & AgentLoop & 32 & \ci{3.339}{0.721} & \ci{193.333}{122.600} & \ci{205.333}{122.767} \\
critical\_wide & Periodic Poll & 32 & \ci{5.605}{0.320} & \ci{2.875}{1.844} & \ci{14.875}{1.844} \\
critical\_wide & ReflexPatch & 32 & \ci{4.917}{0.892} & \ci{11.774}{13.307} & \ci{21.742}{10.936} \\
critical\_wide & EP & 32 & \ci{5.217}{0.691} & \ci{6.387}{8.065} & \ci{19.484}{8.742} \\
\midrule
soft\_keepwarm & AgentLoop & 32 & \ci{3.884}{0.836} & \ci{287.556}{92.973} & \ci{299.556}{93.417} \\
soft\_keepwarm & Periodic Poll & 32 & \ci{5.791}{0.355} & \ci{6.774}{3.452} & \ci{20.774}{3.871} \\
soft\_keepwarm & ReflexPatch & 32 & \ci{5.447}{0.314} & \ci{3.406}{3.938} & \ci{15.406}{4.078} \\
soft\_keepwarm & EP & 32 & \ci{5.211}{0.449} & \ci{0.000}{0.000} & \ci{13.103}{1.655} \\
\bottomrule
\end{tabular}
\end{adjustbox}
\end{table}

% Suggested label: \label{tab:embodied-one-dish-family}

Table~\ref{tab:embodied-one-dish-family} breaks down the one-dish results by dish family. AgentLoop
suffers large visibility and finish delays across all families because cooking signals are surfaced
only after tutoring completes. EP performs best on the critical-medium and critical-narrow families,
where early visibility is most valuable. In critical-wide and soft-keepwarm cases, the differences are
smaller and depend more on whether the controller chooses to defer or respond immediately, because
wider windows reduce the pure latency penalty.

\paragraph{Multi-dish setting.}
The multi-dish-multi-tutor setting is the main selective scheduling benchmark. It contains multiple
dishes and multiple tutoring problems, with overlapping cooking signals and repeated opportunities
to interrupt and resume tutoring. We sweep \(K \in \{3,4,5\}\) dishes, \(M \in \{2,3\}\) tutor
problems, and stove capacity \(S_{\mathrm{stove}} \in \{1,2\}\), yielding \(12\) structural cells.
For each cell, we instantiate three randomized case families and two seeds \((7,11)\), for a total
of \(12 \times 3 \times 2 = 72\) cases.

The three case families stress different scheduling mechanisms.
\texttt{soft\_vs\_critical\_conflict} places an early soft timer before a critical whistle that follows
\(8\)--\(28\) ticks later, together with later background dishes.
\texttt{staggered\_criticals} places two critical whistles \(10\)--\(35\) ticks apart, with additional
background soft or wide-critical dishes.
\texttt{resume\_pressure} places an early critical dish, then a soft dish, then a later critical dish,
explicitly creating repeated interrupt--resume opportunities.

Tutor families are sampled from the same generic mathematics pool as in the one-dish setting, with
case-family-dependent anchors. \texttt{soft\_vs\_critical\_conflict} begins from short and long tutor
problems, \texttt{staggered\_criticals} begins from medium and long tutor problems, and
\texttt{resume\_pressure} begins from long and medium tutor problems, with additional families sampled
when \(M=3\). The runtime configuration matches the one-dish setting except that
\(\texttt{max\_controller\_calls}=60\). In particular, \(\texttt{polling\_interval}=40\), no default
tutor-token budget or tutor-segment cap is imposed, and whistle dishes receive Gaussian signal jitter
with standard deviation \(3.0\).

% Existing Table 18 goes here.
\begin{table}[ht]
\centering
\caption{Multi-dish-multi-tutor embodied scheduling results. Cells report mean $\pm$ half-width of the bootstrap 95\% confidence interval.}
\label{tab:embodied-multi-dish-aggregate}
\begin{adjustbox}{max width=\linewidth}
\begin{tabular}{lrrrrrrrrr}
\toprule
\textbf{Method} & \textbf{$n$} & \textbf{Utility $\uparrow$} & \textbf{Critical succ. $\uparrow$} & \textbf{Soft quality $\uparrow$} & \textbf{Vis. delay $\downarrow$} & \textbf{Finish delay $\downarrow$} & \textbf{Priority} & \textbf{Soft defer} & \textbf{Resume delay} \\
\midrule
AgentLoop & 72 & \ci{19.671}{1.290} & \ci{0.745}{0.068} & \ci{0.833}{0.079} & \ci{74.108}{18.466} & \ci{91.691}{18.819} & \ci{0.870}{0.131} & \ci{0.119}{0.046} & \ci{146.210}{31.160} \\
Periodic Poll & 72 & \ci{18.981}{1.235} & \ci{0.718}{0.076} & \ci{0.971}{0.032} & \ci{17.987}{5.877} & \ci{37.625}{8.398} & \ci{0.857}{0.215} & \ci{0.084}{0.039} & \ci{83.651}{18.810} \\
ReflexPatch & 72 & \ci{21.202}{1.009} & \ci{0.913}{0.051} & \ci{0.952}{0.046} & \ci{17.897}{9.010} & \ci{28.568}{8.614} & -- & \ci{0.069}{0.063} & \ci{93.647}{22.277} \\
EP & 72 & \ci{21.288}{0.852} & \ci{0.936}{0.043} & \ci{0.957}{0.040} & \ci{9.693}{6.548} & \ci{28.537}{8.157} & \ci{0.722}{0.195} & \ci{0.098}{0.044} & \ci{88.605}{15.303} \\
\bottomrule
\end{tabular}
\end{adjustbox}
\end{table}
% Suggested label: \label{tab:embodied-multi-dish-aggregate}

The aggregate multi-dish results in Table~\ref{tab:embodied-multi-dish-aggregate} show that EP obtains
the highest utility and critical-dish success, while ReflexPatch is very close in utility and finish
delay. This comparison should be interpreted carefully: ReflexPatch is an engine-assisted reference,
because the runtime automatically applies a default cooking response before the controller can choose.
EP instead tests whether the controller can use immediate event visibility to make its own scheduling
decisions. The small gap between EP and ReflexPatch suggests that event-time visibility plus
agent-side control can approach a hard-coded reflex reference in this setting.

% Existing Table 19 goes here.
\begin{table}[ht]
\centering
\caption{Multi-dish results by case family. Cells report mean $\pm$ half-width of the bootstrap 95\% confidence interval.}
\label{tab:embodied-multi-dish-family}
\begin{adjustbox}{max width=\linewidth}
\begin{tabular}{llrrrrrr}
\toprule
\textbf{Family} & \textbf{Method} & \textbf{$n$} & \textbf{Utility $\uparrow$} & \textbf{Critical succ. $\uparrow$} & \textbf{Soft quality $\uparrow$} & \textbf{Soft defer} & \textbf{Resume delay} \\
\midrule
resume\_pressure & AgentLoop & 24 & \ci{21.103}{2.527} & \ci{0.785}{0.108} & \ci{0.917}{0.104} & \ci{0.111}{0.080} & \ci{155.136}{55.870} \\
resume\_pressure & Periodic Poll & 24 & \ci{19.133}{2.390} & \ci{0.715}{0.116} & \ci{0.958}{0.063} & \ci{0.097}{0.080} & \ci{96.717}{33.321} \\
resume\_pressure & ReflexPatch & 24 & \ci{22.226}{1.489} & \ci{0.965}{0.045} & \ci{1.000}{0.000} & \ci{0.000}{0.000} & \ci{79.465}{20.468} \\
resume\_pressure & EP & 24 & \ci{21.784}{1.284} & \ci{0.979}{0.031} & \ci{0.938}{0.084} & \ci{0.146}{0.094} & \ci{88.335}{25.339} \\
\midrule
soft\_vs\_critical\_conflict & AgentLoop & 24 & \ci{18.243}{1.455} & \ci{0.674}{0.112} & \ci{0.806}{0.122} & \ci{0.172}{0.083} & \ci{123.927}{43.707} \\
soft\_vs\_critical\_conflict & Periodic Poll & 24 & \ci{20.351}{1.245} & \ci{0.847}{0.104} & \ci{0.979}{0.031} & \ci{0.090}{0.066} & \ci{63.421}{9.035} \\
soft\_vs\_critical\_conflict & ReflexPatch & 24 & \ci{20.572}{1.815} & \ci{0.944}{0.063} & \ci{0.903}{0.097} & \ci{0.208}{0.167} & \ci{64.033}{11.049} \\
soft\_vs\_critical\_conflict & EP & 24 & \ci{21.078}{1.591} & \ci{0.958}{0.052} & \ci{0.938}{0.063} & \ci{0.076}{0.063} & \ci{70.010}{11.129} \\
\midrule
staggered\_criticals & AgentLoop & 24 & \ci{19.668}{2.296} & \ci{0.778}{0.125} & \ci{0.773}{0.159} & \ci{0.073}{0.063} & \ci{159.567}{56.769} \\
staggered\_criticals & Periodic Poll & 24 & \ci{17.460}{2.502} & \ci{0.590}{0.142} & \ci{0.977}{0.034} & \ci{0.066}{0.059} & \ci{90.815}{41.924} \\
staggered\_criticals & ReflexPatch & 24 & \ci{20.809}{1.922} & \ci{0.830}{0.124} & \ci{0.955}{0.068} & \ci{0.000}{0.000} & \ci{137.442}{58.092} \\
staggered\_criticals & EP & 24 & \ci{21.003}{1.566} & \ci{0.872}{0.101} & \ci{1.000}{0.000} & \ci{0.073}{0.063} & \ci{107.469}{34.424} \\
\bottomrule
\end{tabular}
\end{adjustbox}
\end{table}
% Suggested label: \label{tab:embodied-multi-dish-family}

Table~\ref{tab:embodied-multi-dish-family} reports results by case family. The main separation comes
from priority-sensitive cases: EP is strongest on \texttt{soft\_vs\_critical\_conflict} and
\texttt{staggered\_criticals}, whereas ReflexPatch is strongest on \texttt{resume\_pressure}. This
pattern is consistent with the qualitative role of the interfaces. ReflexPatch reacts quickly by
construction, but EP preserves model choice when multiple pending dishes compete for priority or when
a soft-dish signal does not require an immediate switch.

Periodic Poll presents an instructive tradeoff. It achieves high average soft quality, but this
comes at the expense of critical-dish timing; its average critical success is substantially below EP.
This indicates that periodic polling is often sufficient to preserve non-urgent soft dishes, but is
less reliable when multiple urgent signals must be reprioritized inside a single shared schedule.

\paragraph{Tutoring continuity and representative trace.}
A possible concern is that event-time interruption could improve kitchen timing only by degrading
the tutoring stream. We therefore evaluate whether the online tutor subsystem resumes cleanly after
interruptions, and we visualize a representative resume-pressure episode.

\begin{figure}[t]
\centering
\includegraphics[width=\linewidth]{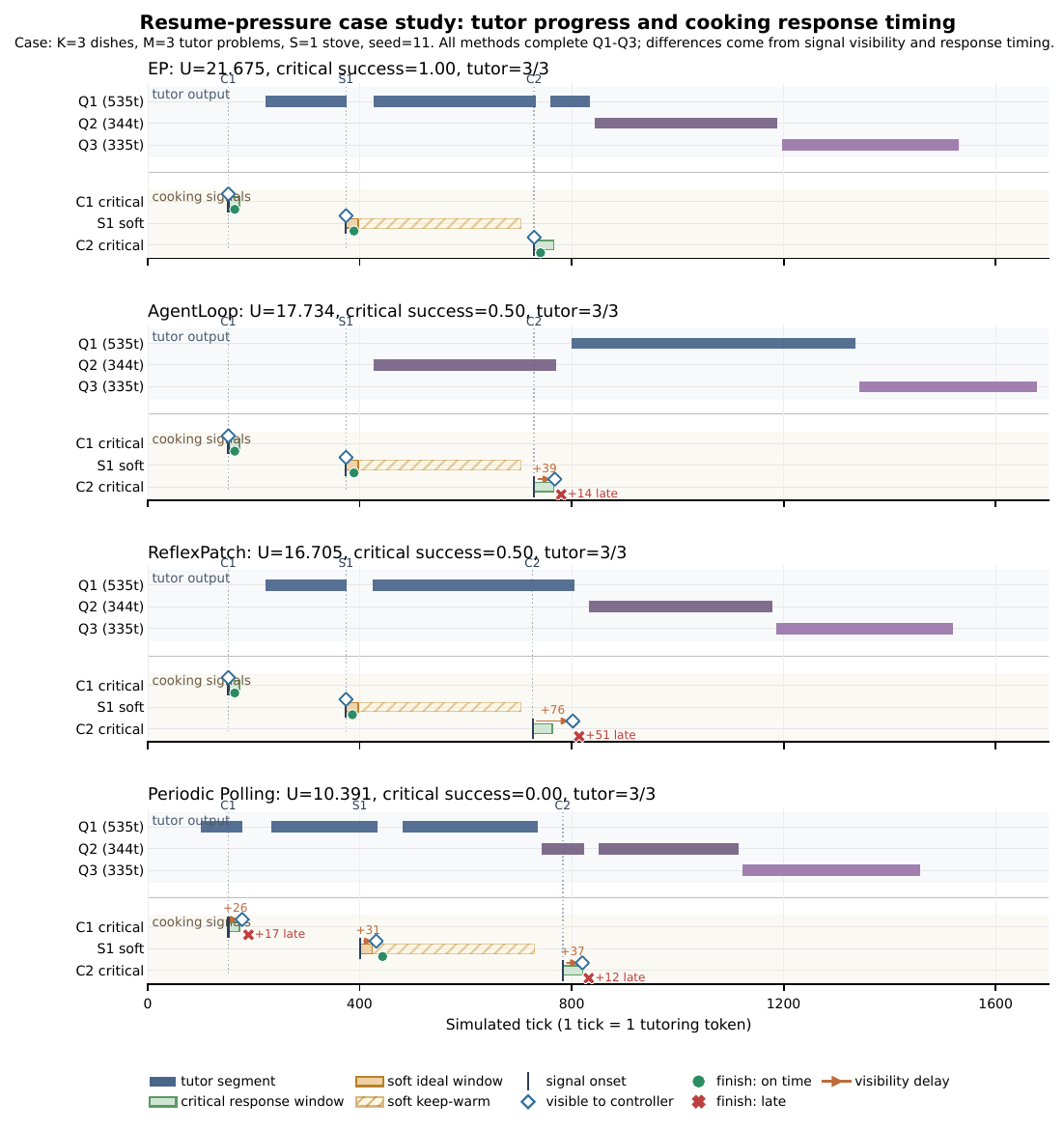}
\caption{Representative episode from the \texttt{resume\_pressure} family with three dishes, three
tutor problems, and one stove slot. The upper lanes show generated tutor segments for Q1--Q3, while
the lower lanes show cooking signals, valid response windows, and finish actions.}
\label{fig:embodied-resume-pressure-trace}
\end{figure}

Figure~\ref{fig:embodied-resume-pressure-trace} shows how the interfaces differ on the same episode.
The difference is whether cooking signals become
visible while tutoring is in progress and who decides the immediate reaction. In this trace, EP
interrupts and resumes tutoring at event time, preserving both critical dishes and the soft dish.
AgentLoop and ReflexPatch complete the same tutoring workload but miss the late critical dish, while
Periodic polling unfortunately misses both critical dishes in this case, because the rigid check intervals happen to fall outside the critical visibility windows.

% Existing Table 20 goes here.
\begin{table}[ht]
\centering
\caption{Tutoring continuity metrics. Cells report mean $\pm$ half-width of the bootstrap 95\% confidence interval.}
\label{tab:embodied-tutor-continuity}
\begin{adjustbox}{max width=\linewidth}
\begin{tabular}{llrrrr}
\toprule
\textbf{Setting} & \textbf{Method} & \textbf{Interrupted segments} & \textbf{Resumed segments} & \textbf{Resume overlap} & \textbf{Restart rate} \\
\midrule
One-dish & AgentLoop & \ci{0.000}{0.000} & \ci{0.000}{0.000} & -- & -- \\
One-dish & Periodic Poll & \ci{0.312}{0.079} & \ci{0.297}{0.082} & \ci{0.000}{0.000} & \ci{0.026}{0.040} \\
One-dish & ReflexPatch & \ci{0.297}{0.078} & \ci{0.297}{0.082} & \ci{0.073}{0.077} & \ci{0.079}{0.092} \\
One-dish & EP & \ci{0.344}{0.082} & \ci{0.344}{0.078} & \ci{0.001}{0.001} & \ci{0.000}{0.000} \\
\midrule
Multi-dish & AgentLoop & \ci{0.000}{0.000} & \ci{0.000}{0.000} & -- & -- \\
Multi-dish & Periodic Poll & \ci{2.014}{0.236} & \ci{1.986}{0.236} & \ci{0.018}{0.019} & \ci{0.017}{0.020} \\
Multi-dish & ReflexPatch & \ci{1.903}{0.285} & \ci{1.903}{0.278} & \ci{0.023}{0.031} & \ci{0.021}{0.028} \\
Multi-dish & EP & \ci{2.028}{0.257} & \ci{2.028}{0.250} & \ci{0.016}{0.018} & \ci{0.015}{0.018} \\
\bottomrule
\end{tabular}
\end{adjustbox}
\end{table}
% Suggested label: \label{tab:embodied-tutor-continuity}

Table~\ref{tab:embodied-tutor-continuity} summarizes tutoring continuity across interrupted and
resumed segments. AgentLoop has no interrupted tutoring segments because tutoring is never interrupted
mid-problem. The other three methods interrupt and resume tutoring segments. Restart rates are low
overall, and EP has the lowest restart rate in both one-dish and multi-dish settings, suggesting that
event-time interruption does not substantially degrade tutoring continuity when the tutor subsystem
is conditioned on the prior transcript and recent shared context.

\section{Time-Aware LLM Training}
\label{app:time_aware_llm_training}

This appendix provides additional details and results for the time-aware LLM
training experiment in Section~6.3. This experiment studies whether an LLM-based
reasoning model can learn to adapt its generation behavior under different
deliberation-time constraints. In EP, action generation is part of the
interaction process and may consume time. For reasoning models, we use the
number of generated tokens as a proxy for deliberation time.

\paragraph{Setup.}

We use Qwen3-8B as the base reasoning model and train it on a difficult subset of DeepMath-103K~\citep{DeepMath103K}. We first use Qwen2.5-32B to assign each problem a difficulty score from 1 to 10, using the math difficulty filtering prompt from Figure~29 of OpenThoughts~\citep{guha2025openthoughts}. We keep problems scored 9 or 10 as the hard subset, yielding approximately 19K training candidates, and reserve a fixed 1K test set. We evaluate the checkpoint at step 90, where the training rewards of all three variants had approximately converged; under our training configuration, this corresponds to using about 12K training examples. For each problem, we randomly assign one of four requested token budgets:
\begin{equation}
    B\in\{1000,2000,4000,8000\}.
\end{equation}
Each budget is paired with a natural-language time instruction:
\begin{itemize}[leftmargin=1.2em, itemsep=1pt, topsep=2pt]
\item $B=1000$: ``Now time is urgent, please give me an answer immediately.''
\item $B=2000$: ``Now there is a little time for you to think, but please think simply and give me an answer.''
\item $B=4000$: ``Now there is more time for you to think, please think carefully and give me an answer.''
\item $B=8000$: ``Now there is a lot of time for you to think, please think carefully and give me an answer.''
\end{itemize}

\paragraph{Training variants.}
We train three variants from the same Qwen3-8B initialization on the same data
for 90 GRPO steps with the same training configuration. The first variant,
\textbf{EP-aware}, uses the EP setting: the reward is 1 only if the final answer
is correct and the generated solution length does not exceed the requested
budget $B$; otherwise the reward is 0. Its maximum generation length during
training is 8000 tokens.

The second variant, \textbf{RLVR-8K}, follows a standard RLVR setting with the
same 8000-token training limit, but removes the budget constraint from the
reward: the model receives reward 1 if the final answer is correct and 0
otherwise. The third variant, \textbf{RLVR-16K}, is identical to RLVR-8K except
that the maximum generation length during training is increased to 16000 tokens.
Thus, comparing EP-aware with RLVR-8K isolates the effect of the EP-aware
budgeted reward, while comparing RLVR-8K with RLVR-16K tests whether the
training-time generation limit itself induces implicit time-aware behavior.

All three variants are trained with full-parameter GRPO from the same Qwen3-8B initialization for 90 steps. We implement full-parameter GRPO training using the open-source \texttt{verl} framework~\citep{sheng2025hybridflow}.\footnote{\url{https://github.com/verl-project/verl}} Each training run uses 16 NVIDIA A100 GPUs with tensor parallel size 4 and pipeline parallel size 2. We use batch size 8 per device, micro-batch size 2, micro-rollout batch size 8, mini-batch size 2 per device, rollout tensor parallel size 4, rollout temperature 1.0, rollout number 10, and data shuffling enabled. All variants use the same training data, train/test split, budget assignments, and GRPO configuration. 

All training runs were conducted on NVIDIA A100 GPUs. Each training run used 16 A100 GPUs with 143{,}771 MiB of memory per GPU. Across the three training variants, the total wall-clock training time was approximately 53 hours, corresponding to roughly 848 A100 GPU-hours.

\paragraph{Evaluation.}
All models are evaluated on the same fixed 1K test set with the same decoding configuration. We use temperature 0.6, top-$p$ 0.95, top-$k$ 20, a maximum generation length of 32768 tokens, and thinking mode enabled. We report \emph{Budgeted Success}, the fraction of examples that are both correct and within the requested token budget; \emph{Accuracy}, the fraction of examples with a correct final answer regardless of length; and \emph{Average Tokens}, the average generated solution length. The full results are reported in Table~\ref{tab:time_aware_llm_training}. For uncertainty estimates, we report bootstrap 95\% confidence intervals over test examples within each model--budget group for Budgeted Success, Accuracy, and Average Tokens.

\begin{table}[t]
\small
\centering
% \scriptsize
\caption{Evaluation with a 32768-token decoding limit. EP-aware training uses a
budgeted reward, while RLVR-8K and RLVR-16K use correctness-only rewards with
8000-token and 16000-token training limits, respectively. We report bootstrap
95\% confidence intervals over test examples for Budgeted Success, Accuracy, and
Average Tokens.}
\label{tab:time_aware_llm_training}
% \resizebox{\linewidth}{!}{%
\begin{adjustbox}{max width=\linewidth}
\begin{tabular}{lrrrrr}
\toprule
\textbf{Model} & \textbf{Budget} & \textbf{Total} & \textbf{Budgeted Succ.} & \textbf{Acc.} & \textbf{Avg. Tokens} \\
\midrule
Qwen3-8B & 1000 & 243 & \ci{0.000}{0.000} & \ci{0.407}{0.062} & \ci{7554.28}{603.08} \\
Qwen3-8B & 2000 & 246 & \ci{0.049}{0.029} & \ci{0.398}{0.061} & \ci{6411.49}{535.93} \\
Qwen3-8B & 4000 & 257 & \ci{0.082}{0.035} & \ci{0.346}{0.058} & \ci{8717.73}{619.59} \\
Qwen3-8B & 8000 & 254 & \ci{0.228}{0.051} & \ci{0.311}{0.059} & \ci{8977.00}{735.46} \\
\midrule
RLVR-16K & 1000 & 243 & \ci{0.037}{0.025} & \ci{0.782}{0.054} & \ci{5120.14}{718.85} \\
RLVR-16K & 2000 & 246 & \ci{0.280}{0.057} & \ci{\textbf{0.805}}{0.049} & \ci{3899.67}{538.11} \\
RLVR-16K & 4000 & 257 & \ci{0.393}{0.062} & \ci{\textbf{0.802}}{0.051} & \ci{5271.25}{397.51} \\
RLVR-16K & 8000 & 254 & \ci{0.669}{0.059} & \ci{0.791}{0.051} & \ci{5599.19}{542.45} \\
\midrule
RLVR-8K & 1000 & 243 & \ci{0.284}{0.058} & \ci{\textbf{0.786}}{0.054} & \ci{1891.52}{193.18} \\
RLVR-8K & 2000 & 246 & \ci{0.557}{0.065} & \ci{0.772}{0.053} & \ci{1754.75}{178.40} \\
RLVR-8K & 4000 & 257 & \ci{0.696}{0.058} & \ci{0.794}{0.051} & \ci{2443.72}{235.46} \\
RLVR-8K & 8000 & 254 & \ci{\textbf{0.795}}{0.051} & \ci{\textbf{0.811}}{0.051} & \ci{2413.02}{243.22} \\
\midrule
EP-aware & 1000 & 243 & \ci{\textbf{0.733}}{0.058} & \ci{0.765}{0.054} & \ci{471.90}{85.87} \\
EP-aware & 2000 & 246 & \ci{\textbf{0.691}}{0.057} & \ci{0.691}{0.057} & \ci{522.23}{84.14} \\
EP-aware & 4000 & 257 & \ci{\textbf{0.755}}{0.055} & \ci{0.763}{0.055} & \ci{1027.77}{99.36} \\
EP-aware & 8000 & 254 & \ci{0.783}{0.051} & \ci{0.783}{0.051} & \ci{1124.06}{124.27} \\
\bottomrule
\end{tabular}%
\end{adjustbox}
\end{table}

\paragraph{Discussion.}
Table~\ref{tab:time_aware_llm_training} shows that RL training improves
unconstrained Accuracy across all trained variants. Both RLVR-8K and RLVR-16K
substantially improve over the base Qwen3-8B model, reaching around 0.77--0.81
Accuracy across budgets. However, high unconstrained Accuracy does not
necessarily imply budget-aware behavior. Under the 1000-token budget, RLVR-8K
achieves 0.786 Accuracy but only 0.284 Budgeted Success, while RLVR-16K achieves
0.782 Accuracy but only 0.037 Budgeted Success. In contrast, EP-aware training
achieves 0.733 Budgeted Success under the same 1000-token budget, showing that
the model learns to satisfy tight time constraints rather than merely solve the
problem correctly.

The comparison between RLVR-8K and RLVR-16K also shows that the maximum
generation length during training has an implicit effect on budgeted behavior.
RLVR-8K produces shorter outputs than RLVR-16K and therefore obtains higher
Budgeted Success under all requested budgets. This suggests that limiting the
training-time generation length can act as a weak implicit time constraint.
Nevertheless, this effect is much weaker than directly optimizing the EP-aware
budgeted objective. EP-aware training achieves substantially higher Budgeted
Success under tight budgets while maintaining competitive Accuracy.

Average token counts further reveal the difference. The correctness-only RLVR
models do not adapt generation length strongly to the requested budget:
RLVR-8K stays around 1.8K--2.4K tokens, while RLVR-16K remains around 3.9K--5.6K
tokens. EP-aware training instead produces much shorter outputs under urgent
budgets and increases length as more time is available, from 420.74 tokens at
the 1000-token budget to 1408.56 tokens at the 8000-token budget. This indicates
that EP-aware training learns a time-adaptive generation policy, supporting the
EP view that action generation time should be modeled as part of the interaction
process rather than treated as an external decoding detail.

\paragraph{Resources and licenses.}
We use DeepMath-103K under the MIT License. The Qwen3-8B and Qwen2.5-32B models are released under the Apache License 2.0. DeepSeek-V3.2 is released under the MIT License. Qwen3.5-Plus is accessed as an Alibaba Cloud Model Studio API model and is governed by the Alibaba Cloud Model Studio Terms of Service. The training implementation uses the open-source \texttt{verl} framework described above.

% --- Internal review checklist (optional; comment out for external sharing) ---
% \newpage
% \input{checklist}

\end{document}